\documentclass[nohyperref]{article}

\usepackage{microtype}
\usepackage{graphicx}
\usepackage{subfigure}
\usepackage{booktabs} 

\usepackage{hyperref}
\usepackage{multirow}

\usepackage[accepted]{icml2022}

\usepackage{amsmath}
\usepackage{amssymb}
\usepackage{mathtools}
\usepackage{amsthm}
\newcommand{\fs}[1]{\scriptsize $\pm$#1}

\usepackage[capitalize,noabbrev]{cleveref}

\theoremstyle{plain}
\newtheorem{theorem}{Theorem}[section]

\theoremstyle{definition}
\newtheorem{definition}[theorem]{Definition}
\newtheorem{assumption}[theorem]{Assumption}
\theoremstyle{remark}
\newtheorem{remark}[theorem]{Remark}

\usepackage[textsize=tiny]{todonotes}
\newcommand{\red}[1]{{\textcolor{red}{#1}}}

\usepackage{amsopn,amsmath,amsthm,amssymb}
\usepackage{wrapfig}
\usepackage{multirow}
 \usepackage{mathtools}

\usepackage{anyfontsize}

\usepackage{color}
\usepackage{tikz}
\usetikzlibrary{arrows,shapes,snakes,automata,backgrounds,fit,petri}
\usepackage{adjustbox}

\newcommand{\RN}[1]{%
	\textup{\lowercase\expandafter{\it \romannumeral#1}}%
}

\makeatletter
\newcommand{\distas}[1]{\mathbin{\overset{#1}{\kern\z@\sim}}}%

\usepackage{enumitem}

\usepackage{algorithm}
\usepackage{algorithmic}

\newcommand{\beq}{\vspace{0mm}\begin{equation}}
\newcommand{\eeq}{\vspace{0mm}\end{equation}}
\newcommand{\beqs}{\vspace{0mm}\begin{eqnarray}}
\newcommand{\eeqs}{\vspace{0mm}\end{eqnarray}}
\newcommand{\barr}{\begin{array}}
\newcommand{\earr}{\end{array}}

\newcommand{\Ocal}{\mathcal{O}}

\DeclareMathOperator{\RR}{\mathbb{R}} 

\ifx\assumption\undefined
\newtheorem{assumption}{Assumption}
\fi

\ifx\definition\undefined
\newtheorem{definition}{Definition}
\fi

\ifx\remark\undefined
\newtheorem{remark}{Remark}
\fi
\newcommand{\norm}[1]{\left\| #1 \right\|}

\newcommand{\Abs}[1]{\left| #1 \right| }

\newcommand{\PVar}[1]{\operatorname{Var}\left[#1\right]}
\newcommand{\PExv}[1]{\mathbf{E}\left[#1\right]}
\usepackage{arydshln}

\icmltitlerunning{A Langevin-like Sampler for Discrete Distributions}

\begin{document}

\twocolumn[
\icmltitle{A Langevin-like Sampler for Discrete Distributions}



\icmlsetsymbol{equal}{*}

\begin{icmlauthorlist}
\icmlauthor{Ruqi Zhang}{yyy}
\icmlauthor{Xingchao Liu}{yyy}
\icmlauthor{Qiang Liu}{yyy}
\end{icmlauthorlist}

\icmlaffiliation{yyy}{The University of Texas at Austin}

\icmlcorrespondingauthor{Ruqi Zhang}{ruqiz@utexas.edu}

\icmlkeywords{Machine Learning, ICML}

\vskip 0.3in
]



\printAffiliationsAndNotice{}  

\begin{abstract}
We propose discrete Langevin proposal (DLP), a simple and scalable gradient-based
proposal for sampling complex high-dimensional discrete distributions. In contrast to Gibbs sampling-based methods, DLP is able to update all coordinates in parallel in a single step and the magnitude of changes is controlled by a stepsize. This allows a cheap and efficient exploration in the space of high-dimensional and strongly correlated variables. We prove the efficiency of DLP by showing that the asymptotic bias of its stationary distribution is zero for log-quadratic distributions, and is small for distributions that are close to being log-quadratic. With DLP, we develop several variants of sampling algorithms, including unadjusted, Metropolis-adjusted, stochastic and preconditioned versions. DLP outperforms many popular alternatives on a wide variety of tasks, including Ising models, restricted Boltzmann machines, deep energy-based models, binary neural networks and language generation.
\end{abstract}

\section{Introduction}
Discrete variables are ubiquitous in machine learning problems ranging from discrete data such as text~\cite{wang2019bert, gu2018non} and genome~\cite{wang2010gibbs}, to discrete models such as low-precision neural networks~\cite{courbariaux2016binarized, peters2018probabilistic}. As data and models become large-scale and complicated, there is an urgent need for efficient sampling algorithms on complex high-dimensional discrete distributions.

Markov Chain Monte Carlo (MCMC) methods are typically used to perform sampling, of which the efficiency is largely affected by the proposal distribution~\citep{brooks2011handbook}. For general discrete distributions, Gibbs sampling is broadly applied, which resamples a variable from
its conditional distribution with the remaining variables
fixed. 
Recently, gradient information has been incorporated in the proposal of Gibbs sampling, leading to a substantial boost to the convergence speed in discrete spaces~\citep{grathwohl2021oops}. However, Gibbs-like proposals often suffer from high-dimensional and highly correlated distributions due to conducting a small update per step. In contrast, proposals in continuous spaces that leverage gradients can usually make large effective moves. One of the most popular methods is the Langevin algorithm~\citep{grenander1994representations,roberts1996exponential,roberts2002langevin}, which drives the sampler towards high probability regions following a Langevin diffusion. Due to its simplicity and efficiency, the Langevin algorithm has been widely used for sampling from complicated high-dimensional continuous distributions in machine learning and deep learning tasks ~\citep{welling2011bayesian, li2016preconditioned, grathwohl2019your, song2019generative}.
Its great success makes us ask: \emph{what is the simplest and most natural analogue of the Langevin algorithm in discrete domains?}

In this paper, we develop such a Langevin-like proposal for discrete distributions, which can update many coordinates of the variable based on one gradient computation. By reforming the proposal from the standard Langevin algorithm, we find that it can be easily adapted to discrete spaces and can be cheaply computed in parallel due to coordinatewise factorization. We call this proposal \emph{discrete Langevin proposal} (DLP). Inheriting from the Langevin algorithm, DLP is able to update all coordinates in a single step in parallel and the magnitude of changes is controlled by a stepsize. Using this proposal, we are able to obtain high-quality samples conveniently on a variety of tasks. We summarize our contributions as the following:
\begin{itemize}
    \item We propose discrete Langevin proposal (DLP), a gradient-based proposal for sampling discrete distributions. DLP is able to update many coordinates in a single step with only one gradient computation. 
    \item We theoretically prove the efficiency of DLP by showing that without a Metropolis-Hastings correction, the asymptotic bias of DLP is zero for log-quadratic distributions, and is small for distributions that are close to being log-quadratic.
    \item With DLP, we develop several variants of sampling algorithms, including unadjusted, Metropolis-adjusted, stochastic and preconditioned versions, indicating the general applicability of DLP for different scenarios. 
    \item We provide extensive experimental results, including Ising models, restricted Boltzmann machines, deep energy-based models, binary Bayesian neural networks and text generation, to demonstrate the superiority of DLP in general settings.
\end{itemize}

\section{Related Work}
\paragraph{Gibbs Sampling-based Methods}
Gibbs sampling is perhaps the \emph{de facto}
method for sampling from general discrete distributions. In each step, it iteratively updates one variable leaving the others unchanged. Updating a block of variables is possible, but typically with an increasing cost along with the increase of the block size. To speed up the convergence of Gibbs sampling in high dimensions, \citet{grathwohl2021oops} uses gradient information to choose which coordinate to update and \citet{titsias2017hamming} introduces auxiliary variables to trade off the number of updated variables in a block for less computation.
However, inheriting from Gibbs sampling, these methods still require a large overhead to make significant changes (e.g. $> 5$ coordinates) to the configuration in one step. 

\paragraph{Locally-Balanced Proposals} Based on the information of a local neighborhood of the current position, locally-balanced proposals have been developed for sampling from discrete distributions~\citep{zanella2020informed}. Later they have been extended to continuous-time Markov processes~\citep{power2019accelerated} and have been tuned via mutual information~\citep{sansone2021lsb}. 
Similar to Gibbs sampling-based methods, this type of proposals is very expensive to construct when the local neighborhood is large, preventing them from making large moves in discrete spaces. A concurrent work~\citep{sun2021path} explores a larger neighborhood by making a sequence of small movements. However, it still only updates one coordinate per gradient computation and the update has to be done in sequence, while on the contrary, our method can update many coordinates based on one gradient computation in parallel.

\paragraph{Continuous Relaxation}
Incorporating gradients in the proposal has been a great success in continuous spaces, such as the Langevin algorithm, Hamiltonian Monte Carlo
(HMC)~\citep{duane1987hybrid,neal2011mcmc} and their variants. To take advantage of this success, continuous relaxation is applied which performs sampling in a continuous space by gradient-based methods and then transforms the collected samples to the original discrete space~\citep{pakman2013auxiliary,nishimura2020discontinuous,han2020stein,zhou2020mixed,jaini2021sampling, zhang2022generative}. The efficiency of continuous relaxation highly depends on the properties of the extended continuous distributions which may be difficult to sample from. As shown in previous work, this type of methods usually does not scale to high dimensional discrete distributions~\citep{grathwohl2021oops}. 

\section{Preliminaries}\label{sec:preliminaries}
We consider sampling from a target distribution
\[\pi(\theta) =\frac{1}{Z} \exp(U(\theta)),~~~~~ \forall \theta \in \Theta, 
\]

where $\theta$ is a $d$-dimensional variable, $\Theta$ is a finite\footnote{We consider finite discrete distributions in the paper. However, our algorithms can be easily extended to infinite distributions. See Appendix~\ref{sec:infinite} for a discussion.} variable domain, $U$ is the energy function, and $Z$ is the normalizing constant for $\pi$ to be a distribution. In this paper, we restrict to a \emph{factorized} domain, that is $\Theta=\prod_{i=1}^d \Theta_i$, and mainly consider $\Theta$ to be $\{0,1\}^d$ or $\{0, 1, \ldots, S-1\}^d$. Additionally, we assume that $U$ can be extended to a differentiable function in $\RR^d$. Many popular models have such natural extensions such as Ising models, Potts models, restricted
Boltzmann machines, and (deep) energy-based models.

\paragraph{Langevin Algorithm}

In continuous spaces, one of the most powerful sampling methods is the Langevin algorithm, which follows a Langevin diffusion to update variables:
\[
\theta' = \theta + \frac{\alpha}{2}\nabla U(\theta) + \sqrt{\alpha}\xi, \hspace{2em}\xi\sim\mathcal{N}\left(0,I_{d\times d}\right),
\]

where $\alpha$ is a stepsize. The gradient helps the sampler to explore high probability regions efficiently.
Generally, computing the gradient and sampling a Gaussian variable can be done cheaply in parallel on CPUs and GPUs.  
As a result, the Langevin algorithm is especially compelling for complex high-dimensional distributions, and has been extensively used in machine learning and deep learning.

\section{Discrete Langevin Proposal}

In this section, we propose discrete Langevin proposal (DLP), a simple counterpart of the Langevin algorithm in discrete domains. 

At the current position $\theta$, the proposal distribution $q(\cdot|\theta)$ produces the next position to move to. As introduced in Section~\ref{sec:preliminaries}, $q(\cdot|\theta)$ of the Langevin algorithm in continuous spaces can be viewed as a Gaussian distribution with mean $\theta +\alpha/2\nabla U(\theta)$ and covariance $\alpha I_{d\times d}$. Obviously we could not use this Gaussian proposal in discrete spaces. However, we notice that by explicitly indicating the spaces where the normalizing constant is computed over, this proposal is essentially applicable to \emph{any} kind of spaces. Specifically, we write out the variable domain $\Theta$ explicitly in the proposal distribution,
\begin{align}
    q(\theta'|\theta) &= \frac{\exp\left(-\frac{1}{2\alpha}\norm{\theta'-\theta-\frac{\alpha}{2}\nabla U(\theta)}_2^2\right)}
{Z_{\Theta}(\theta)},
\label{eq:proposal}
\end{align}
where the normalizing constant is integrated (continuous) or summed (discrete) over $\Theta$ (we use sum below)
\begin{align*}
    Z_{\Theta}(\theta) &= \sum_{\theta'\in\Theta}\exp\left(-\frac{1}{2\alpha}\norm{\theta'-\theta-\frac{\alpha}{2}\nabla U(\theta)}_2^2\right).
\end{align*}
Here, $\Theta$ can be any space without affecting $q$ being a valid proposal. As a special case, when $\Theta=\RR^d$, it follows that $Z_{\RR^d}(\theta)= (2\pi\alpha)^{d/2}$ and recovers the Gaussian proposal in the standard Langevin algorithm. When $\Theta$ is a discrete space, we naturally obtain a gradient-based proposal for discrete variables. 

Computing the sum over the full space in $Z_{\Theta}(\theta)$ is generally very expensive, for example, the cost is $\Ocal(S^d)$ for $\Theta=\{0, 1, \ldots, S-1\}^d$. This is why previous methods often restrict their proposals to a small neighborhood. A key feature of the proposal in Equation~\eqref{eq:proposal} is that it can be factorized \emph{coordinatewisely}. To see this, we write Equation~\eqref{eq:proposal} as $q(\theta'|\theta) =\prod_{i=1}^{d} q_i(\theta_i'|\theta)$, where $q_i(\theta_i'|\theta)$ is a simple categorical distribution of form:
\begin{align}
   \text{Categorical}\Big(\text{Softmax}\Big(\frac{1}{2}\nabla U(\theta)_i(\theta_i'-\theta_i)
    -\frac{(\theta_i'-\theta_i)^2}{2\alpha}\Big)\Big), \label{eq:proposal2}
\end{align}
with $\theta'_i\in\Theta_i$ (note that the equation does not contain the term $(\alpha/2 \nabla U(\theta)_i)^2$ because it is independent of $\theta'$ and will not affect the softmax result). Combining with coordinatewise factorized domain $\Theta$, the above proposal enables us to update each coordinate in parallel after computing the gradient $\nabla U(\theta)$. The cost of gradient computation is also $\Ocal(d)$, therefore,
the overall cost of constructing this proposal depends linearly rather than exponentially on $d$. This allows the sampler to explore the full space with the gradient information without paying a prohibitive cost. 

We denote the proposal in Equation~\eqref{eq:proposal2} as \emph{Discrete Langevin Proposal} (DLP). DLP can be used with or without a Metropolis-Hastings (MH) step~\citep{metropolis1953equation,hastings1970monte}, which is usually combined with proposals to make the Markov chain reversible. Specifically, after generating the next position $\theta'$ from a distribution $q(\cdot|\theta)$, the MH step accepts it with probability
\begin{align}
        \min\left(1,\exp\left(U(\theta')-U(\theta)\right) \frac{q(\theta|\theta')}{q(\theta'|\theta)}\right).\label{eq:mh}
\end{align}

By rejecting some of the proposed positions, the Markov chain is guaranteed to converge asymptotically to the target distribution.

We outline the sampling algorithms using DLP in Algorithm~\ref{alg:dlp}. We call DLP without the MH step as \emph{discrete unadjusted Langevin algorithm} (DULA) and DLP with the MH step as \emph{discrete Metropolis-adjusted Langevin algorithm} (DMALA). Similar to MALA and ULA in continuous spaces~\citep{grenander1994representations,roberts2002langevin}, DMALA contains two gradient computations and two function evaluations and is guaranteed to converge to the target distribution, while DULA may have asymptotic bias, but only requires one gradient computation, which is especially valuable when performing the MH step is expensive such as in large-scale Bayesian inference~\citep{welling2011bayesian,durmus2019high}.

\paragraph{Connection to Locally-Balanced Proposals}
\citet{zanella2020informed} has developed a class of locally-balanced proposals that can be used in both discrete and continuous spaces.
One of the locally-balanced proposals is defined as 
\begin{align*}
    r(\theta'|\theta)\propto\exp\left(\frac{1}{2} U(\theta')-\frac{1}{2} U(\theta)-\frac{\norm{\theta'-\theta}^2}{2\alpha}\right),
\end{align*}
where $\theta'\in\Theta$.
DLP can be viewed as a first-order Taylor series approximation to $r(\theta'|\theta)$ using
\[
U(x)-U(\theta)\approx \nabla U(\theta)^{\intercal}(x-\theta),~~ \forall~ x\in\Theta.
\]
\citet{zanella2020informed} discussed the connection between their proposals and Metropolis-adjusted Langevin algorithm (MALA) in continuous spaces but did not explore it in discrete spaces. \citet{grathwohl2021oops} uses a similar Taylor series approximation for another locally-balanced proposal, 
\begin{align}
    r'(\theta'|\theta)\propto\exp\left(\frac{1}{2} U(\theta')-\frac{1}{2} U(\theta)\right),\label{eq:local-balanced}
\end{align}
where $\theta'$ belongs to a hamming ball centered at $\theta$ with window size 1. Like Gibbs sampling, their proposal only updates one coordinate per step. They also propose an extension of their method to update $X$ coordinates per step, but with $X$ times gradient computations. See Appendix~\ref{sec:stepsize-term} for a discussion on Taylor series approximation for Equation~\eqref{eq:local-balanced} without window sizes.

Beyond previous works, we carefully investigate the properties of the Langevin-like proposal in Equation~\eqref{eq:proposal2} in discrete spaces, providing both convergence analysis and extensive empirical demonstration. We find this simple approach explores the discrete structure surprisingly well, leading to a substantial improvement on a range of tasks. 

\begin{algorithm}[t]
  \caption{Samplers with Discrete Langevin Proposal (DULA and DMALA).}
  \begin{algorithmic}
    \label{alg:dlp}
    \STATE \textbf{given:} Stepsize $\alpha$.
      \LOOP
      \FOR[{{\color{blue} Can be done in parallel}}]{$i = 1:d$}
      \STATE \textbf{construct} $q_i(\cdot|\theta)$ as in Equation~\eqref{eq:proposal2}
      \STATE \textbf{sample} $\theta'_i \sim q_i(\cdot|\theta)$
    \ENDFOR
    \vspace{0.5em}
    \STATE $\triangleright$ Optionally, do the MH step
    \STATE \textbf{compute} $q(\theta'|\theta) = \prod_i q_i(\theta'_i|\theta)$ 
    \STATE\hspace{4em} and $q(\theta|\theta') = \prod_i q_i(\theta_i|\theta')$
    \STATE \textbf{set} $\theta \leftarrow \theta'$ with probability in Equation~\eqref{eq:mh}
    
    \ENDLOOP
    \STATE \textbf{output}: samples $\{\theta_k\}$
    
  \end{algorithmic}
\end{algorithm}

\section{Convergence Analysis for DULA}\label{sec:convergence}
In the previous section, we showed that DLP is a convenient gradient-based proposal for discrete distributions. However, the effectiveness of a proposal also depends on how close its underlying stationary distribution is to the target distribution. Because if it is far, even if using the MH step to correct the bias, the acceptance probability will be very low. In this section, we provide an asymptotic convergence analysis for DULA (i.e. the sampler using DLP without the MH step). Specifically, we first prove in Section~\ref{sec:log-qudratic} that when the stepsize $\alpha\rightarrow 0$, the asymptotic bias of DULA is zero for log-quadratic distributions, which are defined as
\begin{align}
    \pi(\theta)\propto \exp\left(\theta^{\intercal} W\theta + b^{\intercal}\theta\right), \theta\in \Theta\label{eq:log-quadratic}
\end{align}
with some constants $W\in \mathbb{R}^{d\times d}$ and $b\in \mathbb{R}^d$. Without loss of generality, we assume $W$ is symmetric (otherwise we can replace $W$ with $(W+W^{\intercal})/2$ for the eigendecomposition). 

Later in Section~\ref{sec:general-dist}, we extend the result to general distributions where we show the asymptotic bias of DULA is small for distributions that are close
to being log-quadratic. 

\subsection{Convergence on Log-Quadratic Distributions}\label{sec:log-qudratic}
We consider a log-quadratic distribution $\pi(\theta)$ as defined in Equation~\eqref{eq:log-quadratic}. This type of distributions appears in common tasks such as Ising models. The following theorem summarizes DULA's asymptotic accuracy for such $\pi$. 

\begin{theorem}\label{thm:log-quadratic}
If the target distribution $\pi$ is log-quadratic as defined in Equation~\eqref{eq:log-quadratic}.
Then the Markov chain following transition $q(\cdot|\theta)$ in Equation~\eqref{eq:proposal2} (i.e. DULA)
is reversible with respect to some distribution $\pi_{\alpha}$ and $\pi_{\alpha}$ converges weakly to $\pi$ as $\alpha\rightarrow 0$. In particular, let $\lambda_{\text{min}}$ be the smallest eigenvalue of $W$, then for any $\alpha>0$,
\[
\norm{\pi_{\alpha} -\pi}_1\le 
     Z\cdot\exp\left(-\frac{1}{2\alpha}-\frac{\lambda_{\text{min}}}{2}\right),
\]

where $Z$ is the normalizing constant of $\pi$.
\end{theorem}
Theorem~\ref{thm:log-quadratic} shows that the asymptotic bias of DULA decreases at a $\Ocal(\exp(-1/(2\alpha))$ rate which vanishes to zero as the stepsize $\alpha\rightarrow 0$. This is similar to the case of the Langevin algorithm in continuous spaces, where it converges asymptotically when the stepsize goes to zero. 

We empirically verify this theorem in Figure~\ref{fig:theory}a. We run DULA with varying stepsizes on a $2$ by $2$ Ising model. For each stepsize, we run the chain long enough to make sure it converged. The results clearly show that the distance between the stationary distribution of DULA and the target distribution decreases as the stepsize decreases. Moreover, the decreasing speed roughly aligns with a function containing $\exp(-1/(2\alpha))$, which demonstrates the convergence rate with respect to $\alpha$ in Theorem~\ref{thm:log-quadratic}.

\subsection{Convergence on General Distributions}\label{sec:general-dist}
To generalize the convergence result from log-quadratic distributions to general distributions, we first note that the stationary distribution of DULA always exists and is unique since its transition matrix is irreducible~\citep{levin2017markov}. Then we want to make sure the bound is tight in the log-quadratic case and derive one that depends on
how much the distribution differs from being log-quadratic.
A natural measure of this difference is the distance of $\nabla U$ to a linear function. Specifically, we assume that $\exists W\in\mathbb{R}^{d\times d}, b\in\mathbb{R}, \epsilon\in \mathbb{R^+}$, such that
\begin{align}
    \norm{\nabla U(\theta) - (2W\theta+b)}_1\le \epsilon, \forall~ \theta\in\Theta.\label{eq:assumption}
\end{align}

Then we have the following theorem for the asymptotic bias of DULA on general distributions.
\begin{theorem}\label{thm:general}
Let $\pi$ be the target distribution and
$\pi'(\theta) = \exp\left(\theta^{\intercal}W\theta + b\theta\right)/Z'$ be the log-quadratic distribution satisfying the assumption in Equation~\eqref{eq:assumption},
then the stationary distribution of DULA satisfies
\begin{align*}
    \norm{\pi_{\alpha} -\pi}_1
    &\le 
     2c_1\left(\exp\left(c_2\epsilon\right)-1\right) + Z'\exp\left(-\frac{1}{2\alpha}-\frac{\lambda_{\text{min}}}{2}\right),
\end{align*}
where $c_1$ is a constant depending on $\pi'$ and $\alpha$; $c_2$ is a constant depending on $\Theta$ and $\max_{\theta,\theta'\in \Theta}\norm{\theta'-\theta}_{\infty}$.
\end{theorem}

The first term in the bound captures the bias induced by the deviation of $\pi$ from being log-quadratic, which decreases in a  $\Ocal(\exp(\epsilon))$ rate. The second term captures the bias by using a non-zero stepsize which directly follows from Theorem~\ref{thm:log-quadratic}. To ensure a satisfying convergence, Theorem~\ref{thm:general} suggests that we should choose a continuous extension for $U$ of which the gradient is close to a linear function.

\paragraph{Example.}
To empirically verify Theorem~\ref{thm:general}, we consider a 1-d distribution $\pi(\theta)\propto \exp\left(a\theta^2+b\theta +2\epsilon\sin(\theta\pi/2)\right)$ where $\theta\in \{-1,1\}$ and $a,b,\epsilon\in \mathbb{R}$. The gradient is $\nabla U(\theta) = 2a\theta + b+\epsilon\pi\cos(\theta\pi/2)$ of which the closeness to a linear function in $\RR$ is controlled by $\epsilon$. From Figure~\ref{fig:theory}b, we can see that when $\epsilon$ decreases, the asymptotic bias of DULA decrease, aligning with Theorem~\ref{thm:general}. 

In fact, the example of $\pi(\theta)\propto \exp\left(2\epsilon\sin(\theta\pi/2)\right)$ with $\theta\in \{-1,1\}$ can be considered as an counterexample for $any$ existing gradient-based proposals in discrete domains.
The gradient on $\{-1,1\}$ is always zero regardless the value of $\epsilon$ whereas the target distribution clearly depends on $\epsilon$. This suggests that we should be careful with the choice of the continuous extension of $U$ since some extensions will not provide useful gradient information to guide the exploration. Our Theorem~\ref{thm:general} provides a guide about how to choose such an extension. 

\begin{figure}[ht]\label{fig:theory}
  \begin{center}
  \begin{tabular}{ccc}
  \hspace{-10pt}
    \includegraphics[width=.24\textwidth]{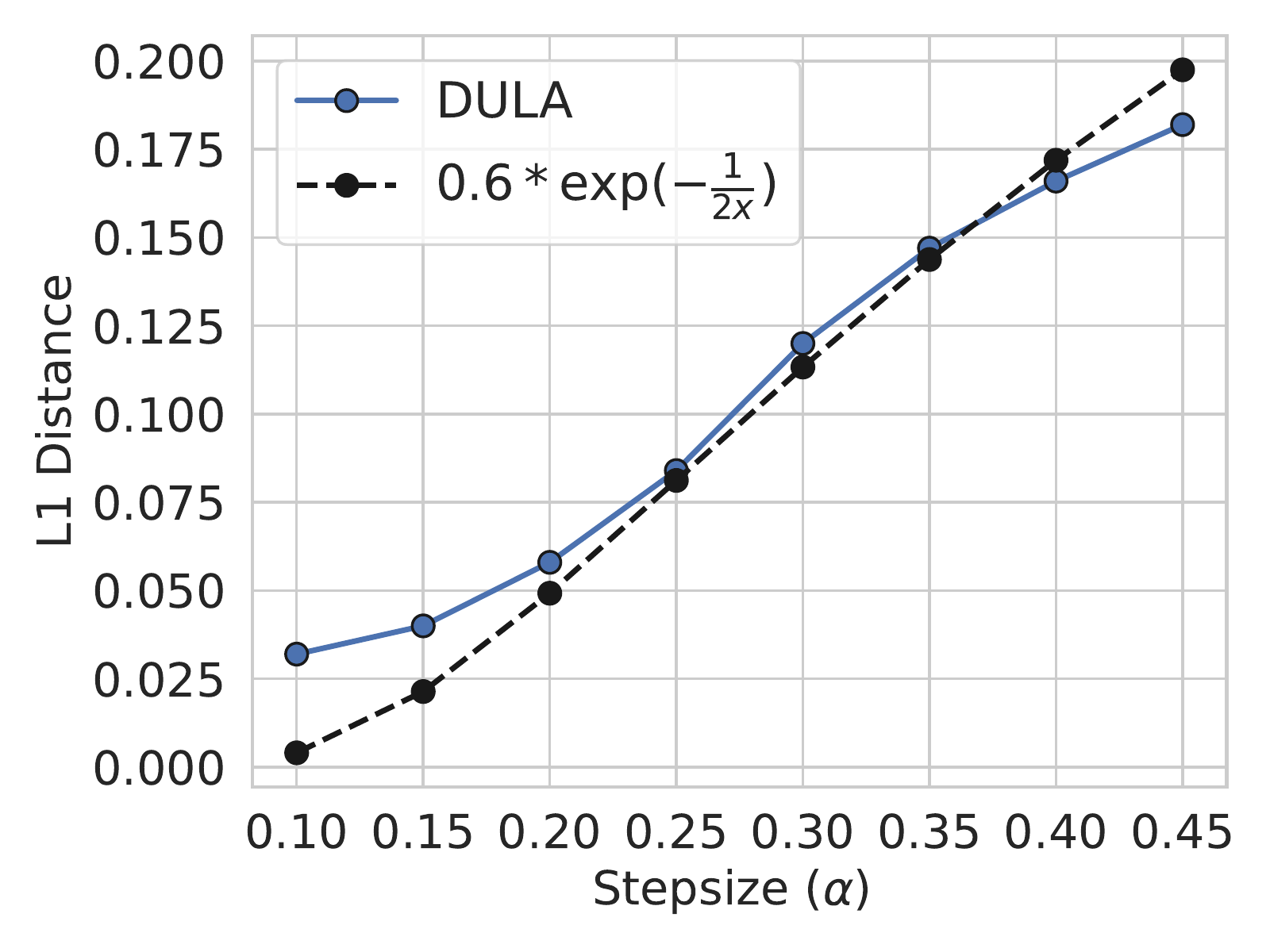} &
    \hspace{-1em}\includegraphics[width=.24\textwidth]{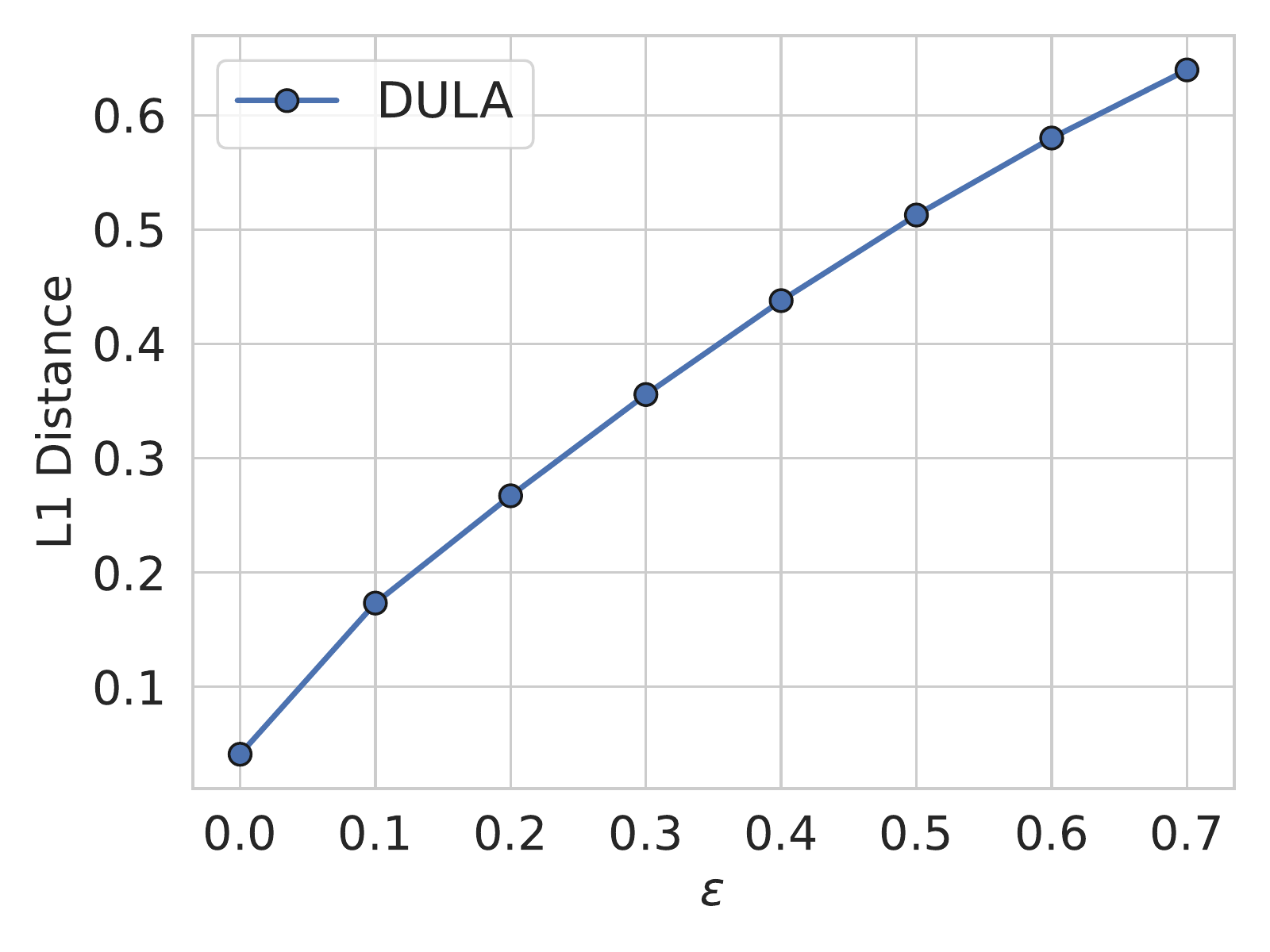}
    \end{tabular}
  \end{center}
  \vspace{-15pt}
    \caption{Empirical verification of theorems. \textbf{Left}: DULA with varying stepsizes on an Ising model. \textbf{Right}: 1-d distribution with varying closeness $\epsilon$ to being log-quadratic.}
    \label{fig:theory}
  \vspace{-10pt}
\end{figure}

\section{Other Variants}\label{sec:variants}

Thanks to the similarity with the standard Langevin algorithm, discrete Langevin proposal can be extended to different usage scenarios following the rich literature of the standard Langevin algorithm. We briefly discuss two such variants in this section.

\paragraph{With Stochastic Gradients}
Similar to Stochastic gradient Langevin dynamics (SGLD)~\citep{welling2011bayesian}, we can replace the full-batch gradient with an unbiased stochastic estimation $\hat{\nabla}U$ in DLP, which will further reduce the cost of our method on large-scale problems. To show the influence of stochastic estimation, we consider a binary domain for simplicity. We further assume that the stochastic gradient has a bounded variance and the norm of the true gradient and the stochastic gradient are bounded. Then we have the following theorem.

\begin{theorem}\label{thm:stochastic}
Let $\Theta=\{0,1\}^d$. We assume that the true gradient $\nabla U$ and the stochastic gradient $\hat{\nabla} U$ satisfy: $\mathbf{E}[\hat{\nabla}U_i]=\nabla U_i$, $\mathbf{E}[\hat{\nabla}U_i]\le \sigma^2$ for some constant $\sigma$;  $|\hat{\nabla}U_i|, \Abs{\nabla U_i} \le L$ for some constant $L$.
Let $q_i$ and $\hat{q}_i$ be the discrete Langevin proposal for the coordinate $i$ using the full-batch gradient and the stochastic gradient respectively, then
\begin{align*}
    \norm{\PExv{\hat{q}_i} - q_i}_1
    &\le 
    2\sigma\cdot\exp\left(-\frac{1}{2\alpha}+L\right).
\end{align*}
\end{theorem}

This suggests that when the variance of the stochastic gradient or the stepsize decreases, the stochastic DLP in expectation will be closer to the full-batch DLP. We test DULA with the stochastic gradient on a binary Bayesian neural network and empirically verify it works well in practice in Appendix~\ref{app:bnn}. 

\paragraph{With Preconditioners} When $\pi$ alters more quickly in some coordinates than others, a single stepsize may result in slow mixing. Under this situation, a preconditiner that adapts the stepsize for different coordinates can help alleviate this problem. We show that it is easy for DLP to incorporate diagonal preconditioners, as long as the coordinatewise factorization still holds. For example, when the preconditioner is constant, that is, we scale each coordinate by a number $g_i$, discrete Langevin proposal becomes 
\begin{align*}
   q_i(\theta_i'|\theta)
    &\propto\exp\left(\frac{1}{2}\nabla U(\theta)_i(\theta_i'-\theta_i)-\frac{(\theta_i-\theta_i')^2}{2\alpha g_i}\right).
\end{align*}
Similar to preconditioners in continuous spaces, $g_i$ adjusts the stepsize for different coordinates considering their variation speed. 
The above proposal is obtained by applying a coordinate transformation to $\theta$ and then transforming the DLP update back to the original space. In this way, the theoretical results in Section~\ref{sec:convergence} directly apply to it. We put more details in Appendix~\ref{sec:preconditioner}.

\section{Experiments}\label{sec:exps}
We conduct a thorough empirical evaluation of discrete Langevin proposal (DLP), comparing to a range of popular baselines, including Gibbs sampling, Gibbs with Gradient (GWG)~\citep{grathwohl2021oops}, Hamming ball (HB)~\citep{titsias2017hamming}---three Gibbs-based approaches; discrete Stein Variational Gradient Descent (DSVGD)~\citep{han2020stein} and relaxed MALA (RMALA)~\citep{grathwohl2021oops}---two continuous relaxation methods; and a locally balanced sampler (LB-1)~\citep{zanella2020informed} which uses the locally-balanced proposal in Equation~\eqref{eq:local-balanced} with window size $1$.
We denote Gibbs-$X$ for Gibbs sampling with a
block-size of $X$, GWG-$X$ for GWG with $X$ indices being modified per step (see D.2 in \citet{grathwohl2021oops} for more details of GWG-$X$), HB-$X$-$Y$ for HB with a block size of $X$ and a hamming ball size of $Y$.
All methods are implemented in \texttt{Pytorch} and we use the official release of code from previous papers when possible. In our implementation, DMALA (i.e. discrete Langevin proposal with an MH step) has a similar cost per step with GWG-$1$ (the main costs for them are one gradient computation and two function evaluations), which is roughly 2.5x of Gibbs-$1$. DULA (i.e. discrete Langevin proposal without an MH step) has a similar cost per step with Gibbs-$1$ (the main cost for DULA is one gradient computation, and for Gibbs-$1$ is $d$ function evaluations). We released the code at \url{https://github.com/ruqizhang/discrete-langevin}.

\subsection{Sampling From Ising Models}
\label{sec:exp:ising}
\begin{figure}[ht]
  \begin{center}
  \begin{tabular}{ccc}
  \hspace{-10pt}
  \includegraphics[width=.23\textwidth]{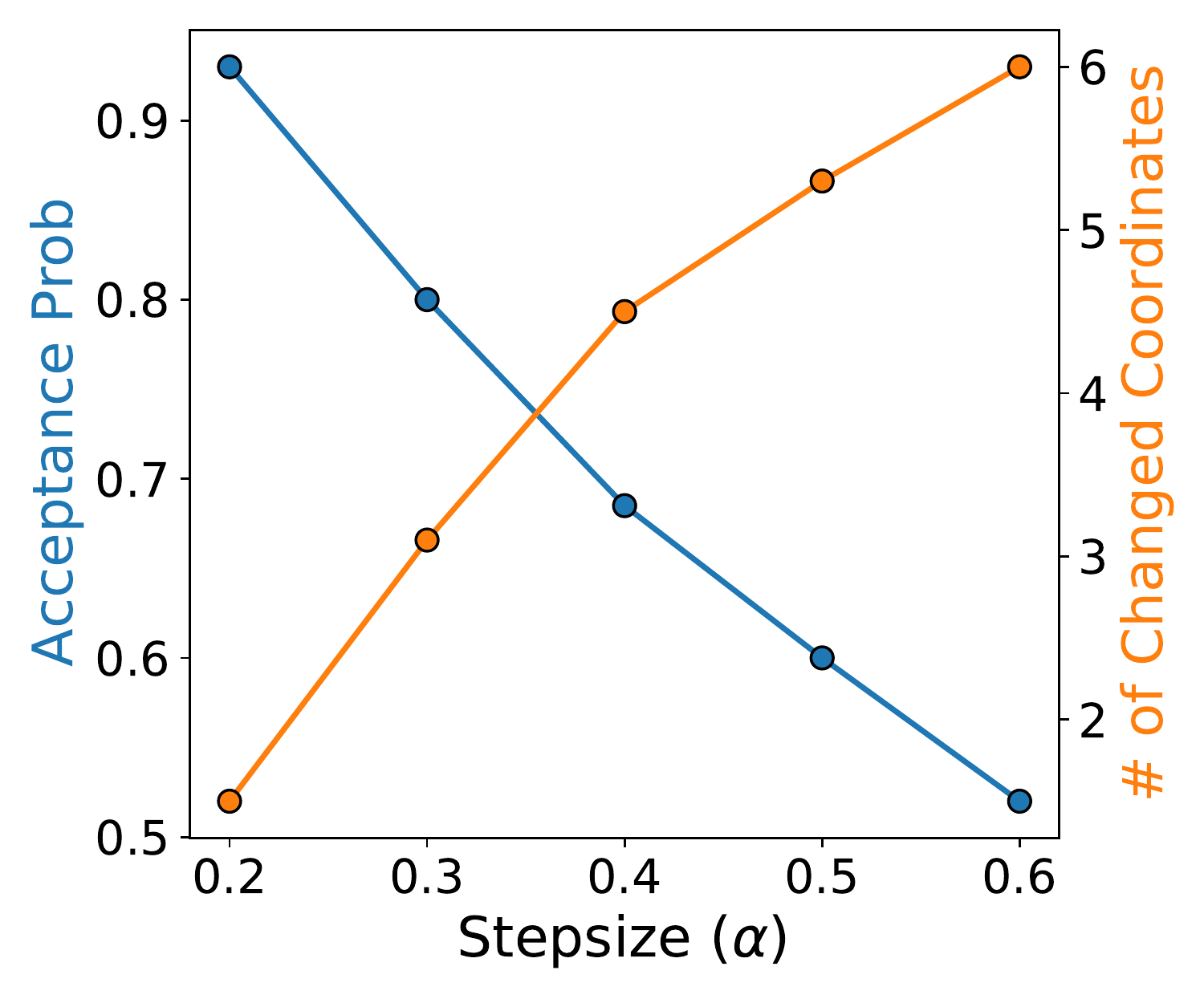}&
  \includegraphics[width=.23\textwidth]{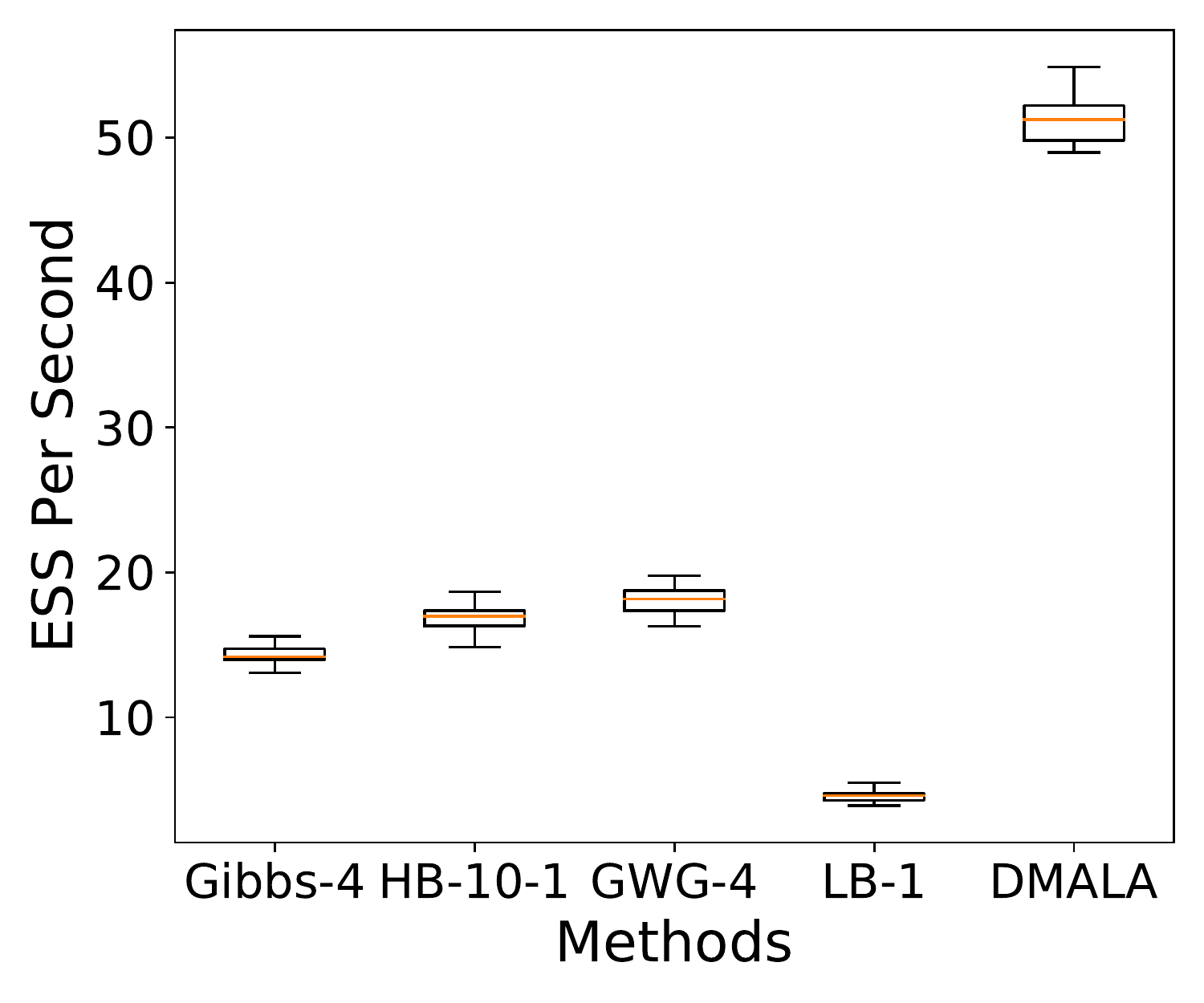}
    \end{tabular}
  \end{center}
  \vspace{-10pt}
    \caption{Ising model sampling results. \textbf{Left}: DLP is able to change many coordinates
while keeping a high acceptance rate. \textbf{Right}: DMALA yields the largest effective sample size (ESS) per second among all the methods compared.}
    \label{fig:ising-ablation}
\end{figure}

\begin{figure}[ht]
  \begin{center}
  \begin{tabular}{ccc}
  \hspace{-10pt}
    \includegraphics[width=.23\textwidth]{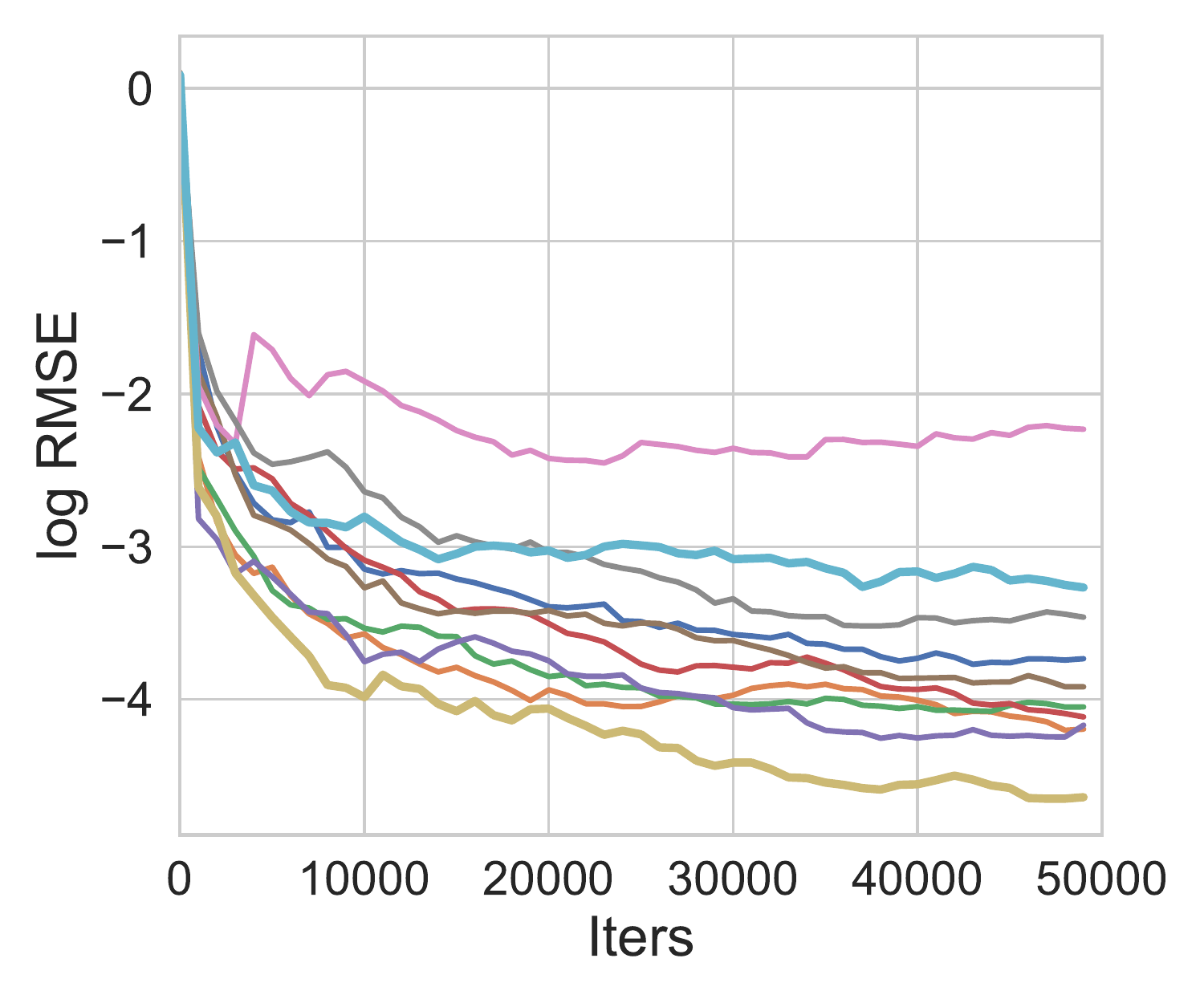} &
    \includegraphics[width=.23\textwidth]{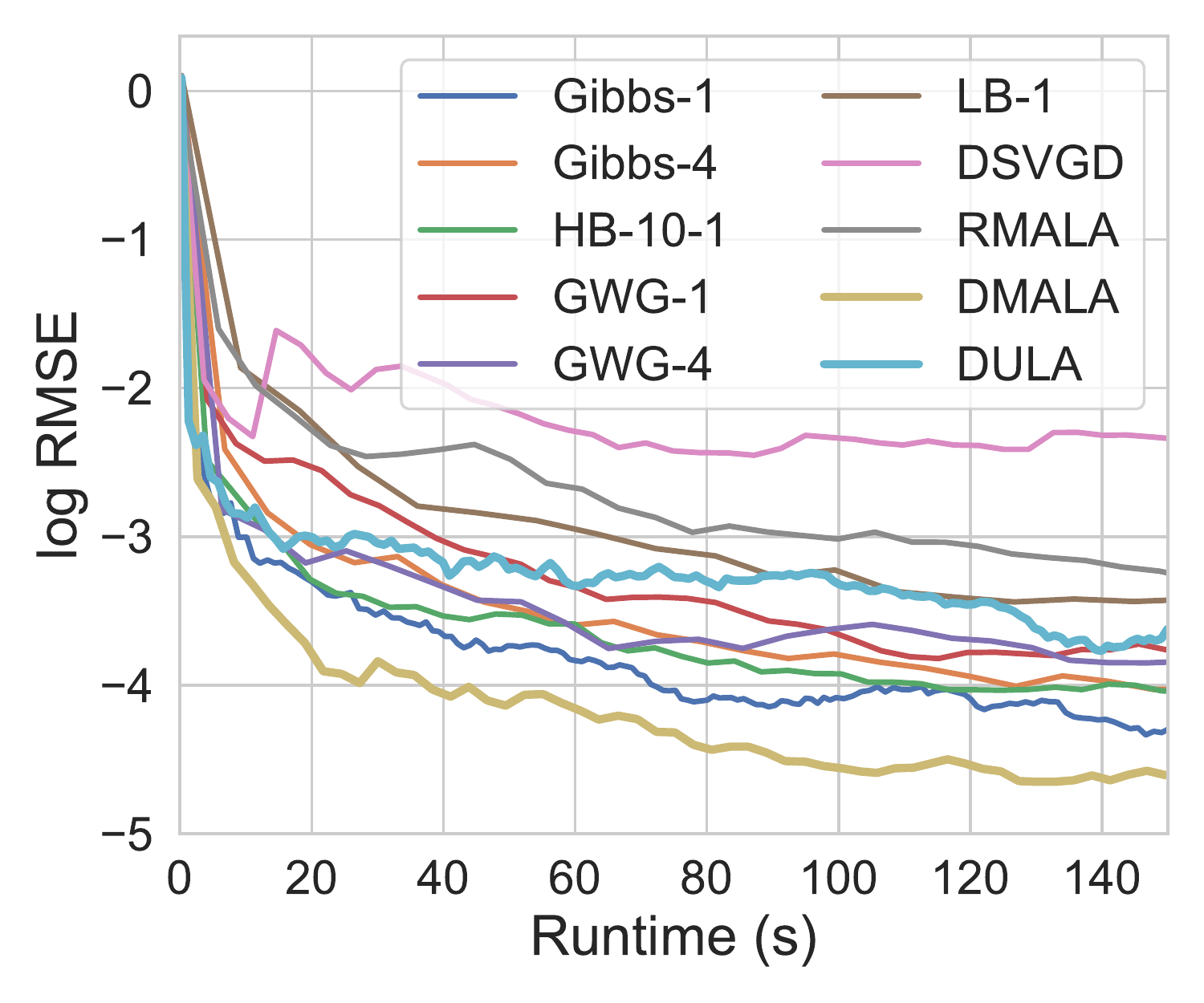} \\
    \end{tabular}
  \end{center}
  \vspace{-10pt}
    \caption{Ising model sampling results. DMALA outperforms the baselines in both number of iterations and running time.}
    \label{fig:ising-sample}
    \vspace{-10pt}
\end{figure}

We consider a $5$ by $5$ lattice Ising model with random variable $\theta \in \{-1,1\}^d$, and $d=5 \times 5 = 25$.  The energy function is
\[
U(\theta) = a\theta^{\intercal}W\theta + b\theta,
\]
where $W$ is a binary adjacency matrix, $a=0.1$ is the connectivity strength and $b=0.2$ is the bias. 
We first show that DLP can change many coordinates in one iteration while still maintaining a high acceptance rate in Figure~\ref{fig:ising-ablation} (Left).
When the stepsize $\alpha=0.6$, on average DMALA can change 6 coordinates in one iteration with an acceptance rate 52\% in the MH step.  
In comparison, GWG-6 (which at most changes 6 coordinates) only has 43\% acceptance rate, not to mention it requires 6x cost of DMALA.
This demonstrates that DLP can make large and effective moves in discrete spaces.
We compare the root-mean-square error (RMSE) between the estimated mean and the true mean in Figure~\ref{fig:ising-sample}. DMALA is the fastest to converge in terms of both runtime and iterations. This demonstrates the importance of (1) using gradient information to explore the space compared to Gibbs and HB; (2) sampling in the original discrete space compared to DSVGD and RMALA; and (3) changing many coordinates in one step compared to LB-1 and GWG-1. GWG-4 underperforms because of a lower acceptance rate than DMALA. DULA can achieve a similar result as LB-1 and GWG-1 but worse than DMALA, indicating that the MH step accelerates the convergence on this task. In Figure~\ref{fig:ising-ablation} (Right), we compare the effective sample size (ESS) per second for exact samplers (i.e. having the target distribution as its stationary distribution). DMALA significantly outperforms other methods, indicating the correlation among its samples is low due to making significant updates in each step. We additionally present results on Ising models with different connectivity strength $a$ in Appendix~\ref{sec:appendix-ising}. 

In what follows, we mainly compare our method with GWG-1 and Gibbs-1, as other methods either could not give reasonable results or are too costly to run.

\subsection{Sampling From Restricted Boltzmann Machines}\label{sec:exp:rbm}
Restricted Boltzmann Machines (RBMs) are generative artificial neural networks, which learn an unnormalized probability over inputs,
\begin{equation*}
    U(\theta) = \sum_i\text{Softplus}( W \theta + a)_i + b ^ {\intercal} \theta, 
\end{equation*}
where $\{W, a, b\}$ are parameters and $\theta \in \{0,1\}^d$. 
Following ~\citet{grathwohl2021oops},
we train $\{W, a, b\}$ with contrastive divergence~\citep{hinton2002training} on the MNIST dataset for one epoch.
We measure the Maximum Mean Discrepancy (MMD) between the obtained samples and those from Block-Gibbs sampling, which utilizes the known structure. 

\begin{figure}[ht]
  \begin{center}
  \begin{tabular}{cc}
    \hspace{-5pt}\includegraphics[width=.23\textwidth]{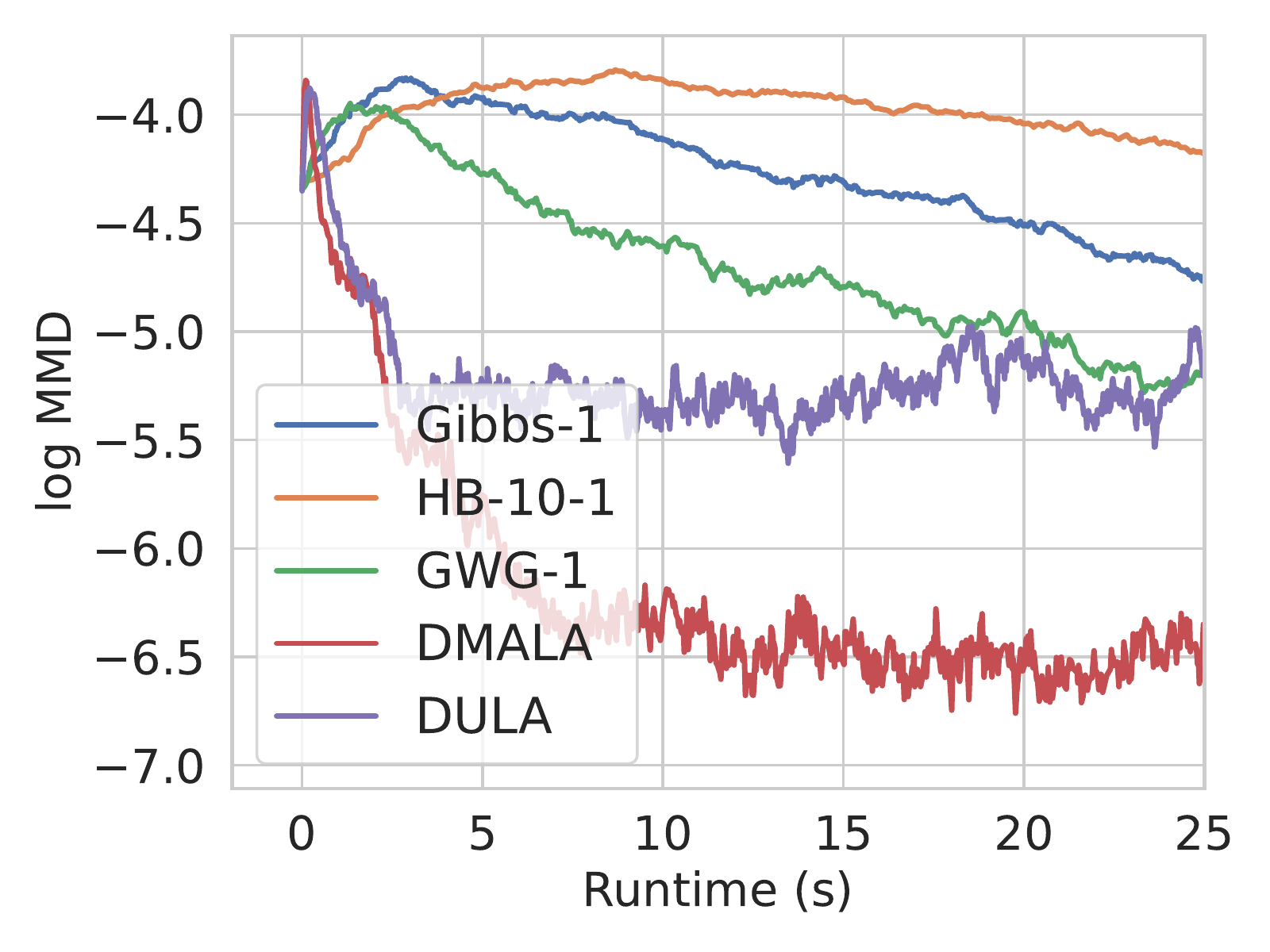} &
    \includegraphics[width=.23\textwidth]{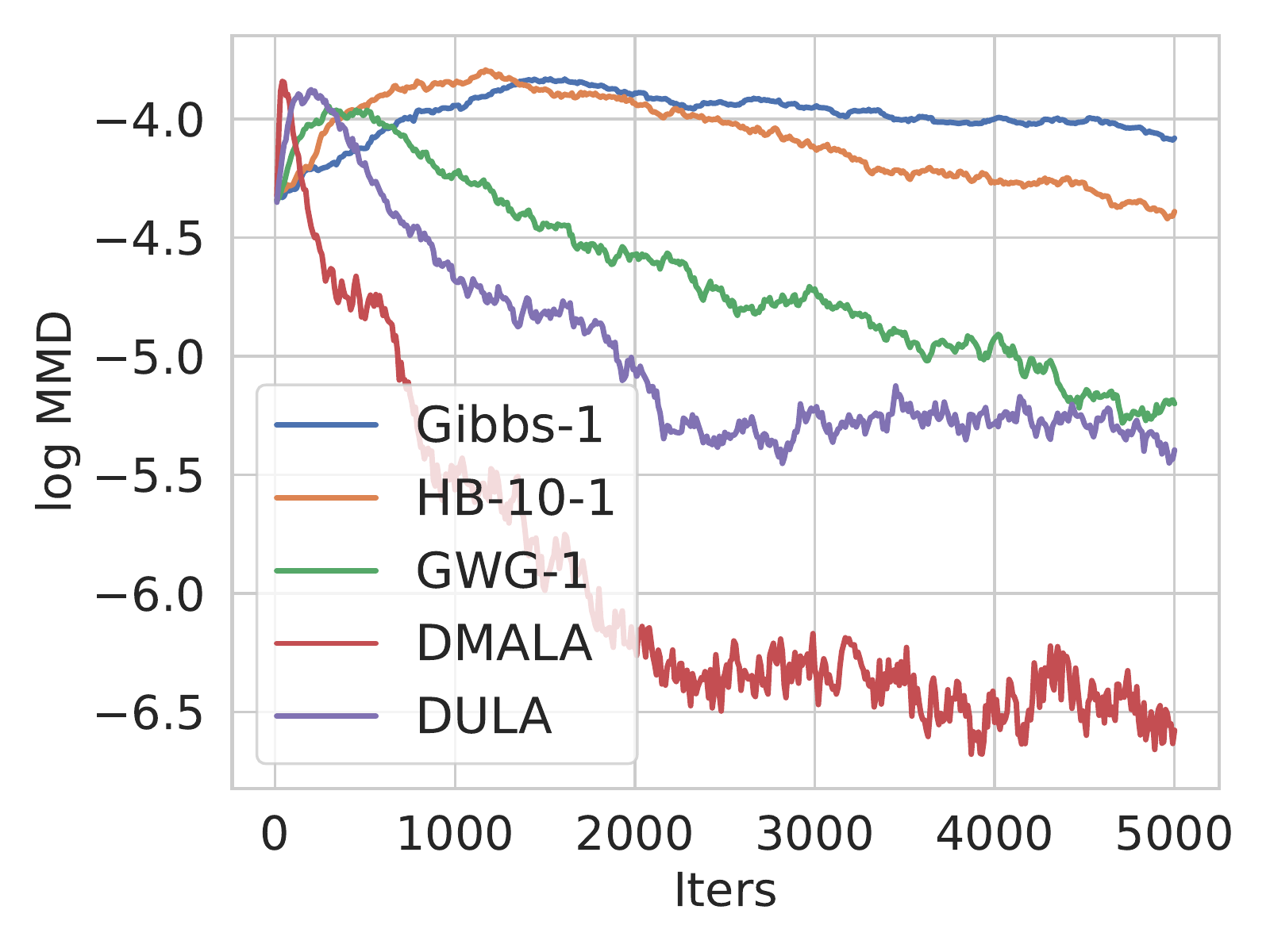}
    \end{tabular}
  \end{center}
  \vspace{-10pt}
    \caption{RBM sampling results. DULA and DMALA converge to the true distribution faster than other methods.}
    \vspace{-10pt}
    \label{fig:rbm-sample}
\end{figure}

\textbf{Results} The results are shown in Figure~\ref{fig:rbm-sample}. Since  DULA and DMALA can update multiple coordinates in a single iteration, they converge remarkably faster than baselines in both iterations and runtime.
Furthermore, DMALA reaches the lowest MMD ($\approx-6.5$) among all the methods after 5,000 iterations, demonstrating the importance of the MH step.
We leave the sampled images in Appendix~\ref{sec:appendix-rbm}.

\begin{figure*}[ht]
  \begin{center}
  \begin{tabular}{ccc}
    \includegraphics[width=.25\textwidth]{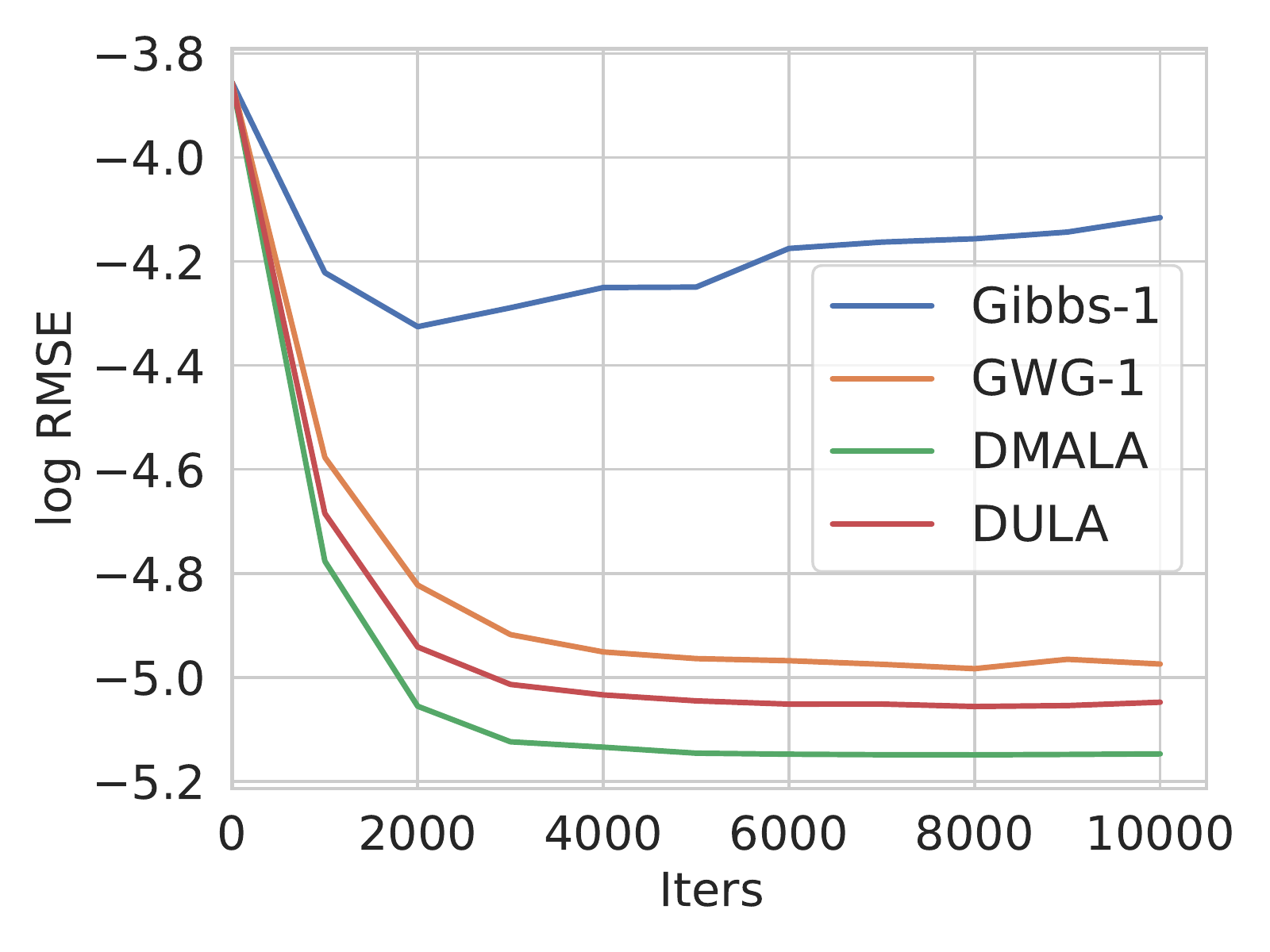} &
    \includegraphics[width=.25\textwidth]{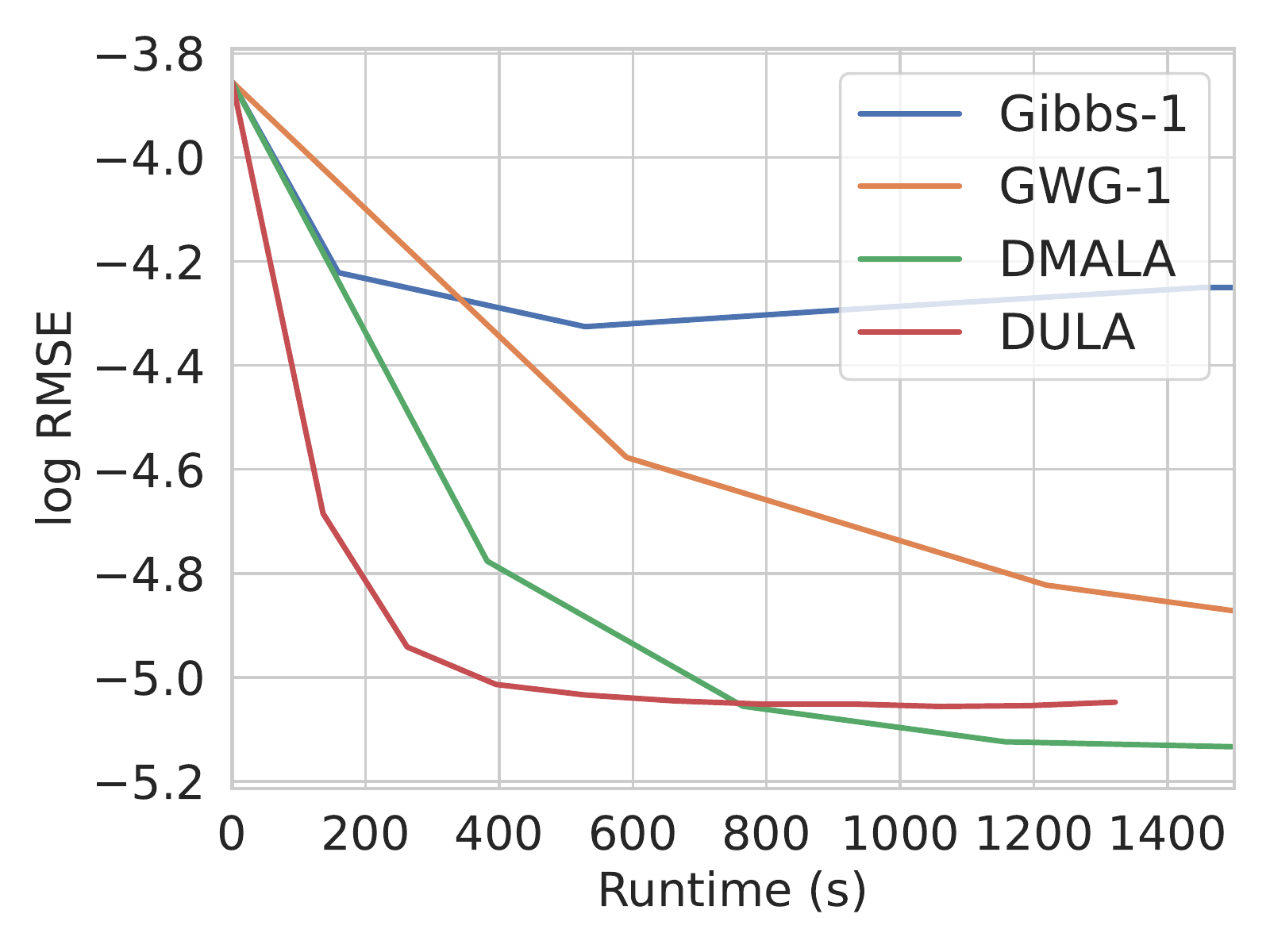}&
    \includegraphics[width=.25\textwidth]{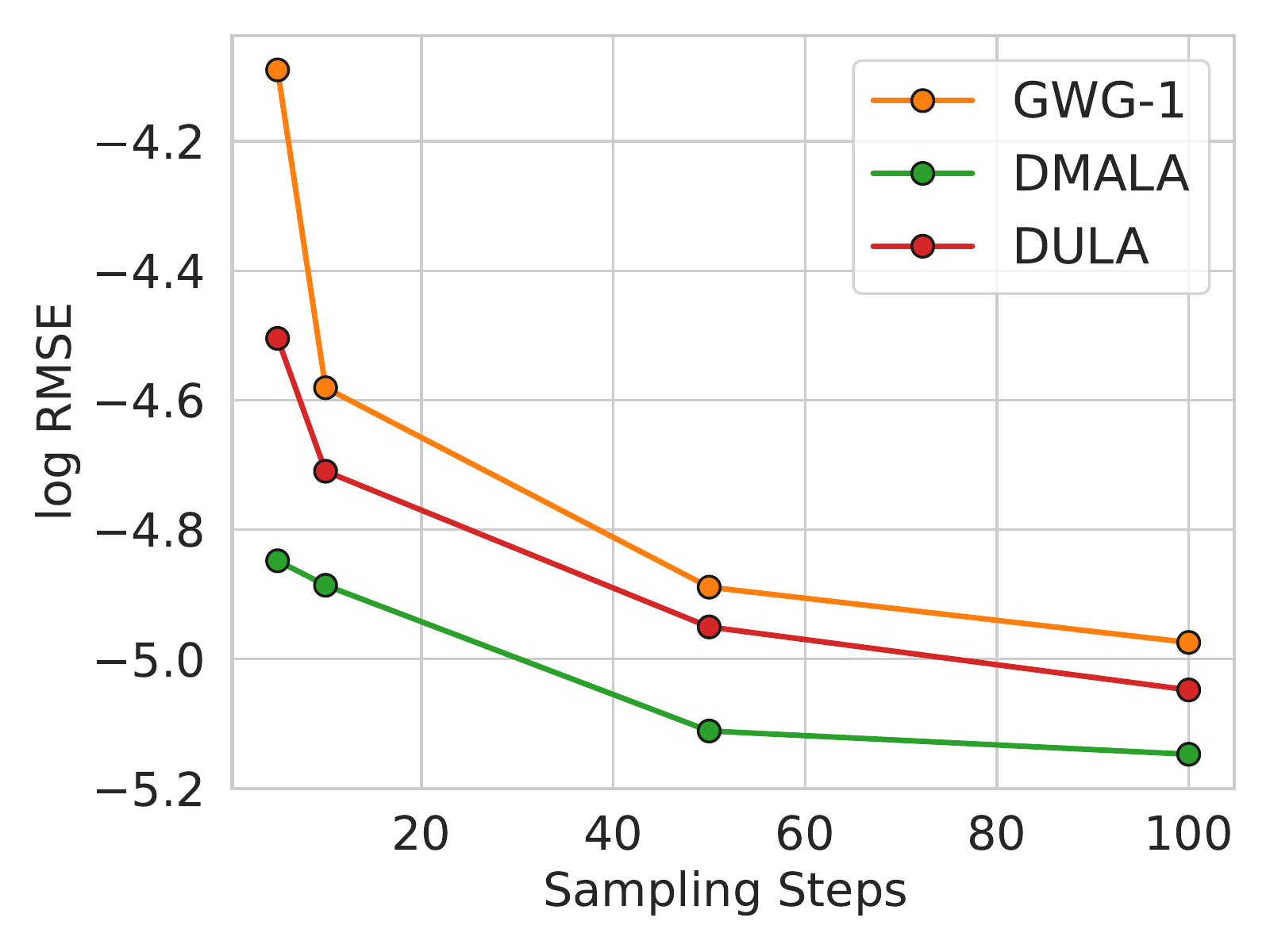} 
    \\
    (a) & (b) & (c)  
    \end{tabular}
  \end{center}
  \vspace{-10pt}
    \caption{Ising model learning results. DULA and DMALA find $W$ in a shorter time than baselines, and the improvement gets larger with smaller number of sampling steps.}
    \label{fig:ising-learn}
\end{figure*}

\begin{table*}[ht]
\centering
\resizebox{0.95\textwidth}{!}{
\begin{tabular}{l|ccc|cc} \hline \hline
 Dataset & VAE (Conv) &EBM (Gibbs) &  EBM (GWG) & EBM (DULA) & EBM (DMALA)  \\ \hline
 Static MNIST  &-82.41   &-117.17  &-80.01  &-80.71  &\textbf{-79.46}  \\
      Dynamic MNIST &-80.40  &-121.19  &-80.51 &-81.29  &\textbf{-79.54}  \\ 
      Omniglot  &-97.65   &-142.06    &-94.72   &-145.68  &\textbf{-91.11} \\
      Caltech Silhouettes &-106.35   &-163.50 &-96.20 &-100.52 &\textbf{-87.82}  \\
\hline \hline
\end{tabular}}
\caption{EBM learning results. We report the log-likelihoods on the test set for different models trained on discrete image datasets.}
\label{tab:ebm}
\end{table*}

\begin{table*}[ht]
    \centering
    \resizebox{0.99\textwidth}{!}{
    \renewcommand\arraystretch{1.2}{
    \begin{tabular}{c|cccc|cccc}
    \hline \hline
        \multirow{2}{*}{Dataset}&  \multicolumn{4}{c|}{Training Log-likelihood ($\uparrow$)} & \multicolumn{4}{c}{Test RMSE ($\downarrow$)} \\ \cline{2-9}
         & Gibbs & GWG & DULA & DMALA & Gibbs & GWG & DULA & DMALA \\ \hline
        COMPAS & -0.4063 \fs{0.0016} & -0.3729 \fs{0.0019} & -0.3590 \fs{0.0003} & \textbf{-0.3394 \fs{0.0016}} & \textbf{0.4718 \fs{0.0022}} & 0.4757 \fs{0.0008} & 0.4879 \fs{0.0024} & 0.4879 \fs{0.0003}  \\
        News & -0.2374 \fs{0.0003} & -0.2368 \fs{0.0003} & \textbf{-0.2339 \fs{0.0003}} & \textbf{-0.2344 \fs{0.0005}} & 0.1048 \fs{0.0025} & 0.1028 \fs{0.0023} & 0.0971 \fs{0.0012} & \textbf{0.0943 \fs{0.0009}} \\
        Adult & -0.4587 \fs{0.0029} & -0.4444 \fs{0.0041} & -0.3287 \fs{0.0027} & \textbf{-0.3190 \fs{0.0019}} & 0.4826 \fs{0.0027} & 0.4709 \fs{0.0034} & 0.3971 \fs{0.0014} & \textbf{0.3931 \fs{0.0019}}  \\ 
        Blog & -0.3829 \fs{0.0036} & -0.3414 \fs{0.0028} & -0.2694 \fs{0.0025} & \textbf{-0.2670 \fs{0.0031}} & 0.4191 \fs{0.0061} & 0.3728 \fs{0.0034} & \textbf{0.3193 \fs{0.0021}} & \textbf{0.3198 \fs{0.0043}} \\
    \hline \hline
    \end{tabular}}}
    \caption{Experiment results with binary Bayesian neural networks on different datasets.}
    \label{tab:bayesbnn}
\end{table*}

\subsection{Learning Energy-based Models}

Energy-based models (EBMs) have achieved great success in various areas in machine learning~\cite{lecun2006tutorial}.
Generally, the density of an energy-based model is defined as
    $p_\theta(x) = \exp(-E_\theta (x)) / Z_\theta$,
where $E_\theta$ is a function parameterized by $\theta$ and $Z_\theta$ is the normalizing constant. Training EBM usually involves maximizing the log-likelihood,
$\mathcal{L}(\theta) \triangleq \mathbb{E}_{x \sim p_{data}} \left[ \log p_\theta (x) \right]$. However, direct optimization needs to compute $Z_\theta$, which is intractable in most scenarios. To deal with it, we usually estimate the gradient of the log-likelihood instead,
\begin{equation*}
    \nabla\mathcal{L}(\theta) 
    = \mathbb{E}_{x \sim p_\theta}\left[ \nabla_\theta E_\theta (x) \right] - \mathbb{E}_{x \sim p_{data}} \left[ \nabla_\theta E_\theta (x) \right].
\end{equation*}
Though the first term is easy to estimate from the data, the second term requires samples from $p_\theta$. Better samplers can improve the training process of $E_\theta$, leading to EBMs with higher performance. 

\subsubsection{Ising models}
As in ~\citet{grathwohl2021oops}, we generate a 25 by 25 Ising model and generate training data by running a Gibbs sampler. In this experiment, $E_\theta$ is an Ising model with learnable parameter $\hat{W}$. We evaluate the samplers by computing RMSE between the estimated $\hat{W}$ and the true $W$.

\textbf{Results}~~~Our results are summarized in Figure~\ref{fig:ising-learn}. In (a), DMALA and DULA always have smaller RMSE than baselines given the same number of iterations. In (b), DMALA and DULA get a log-RMSE of $-5.0$ in $800$s, while the baseline methods fail to reach $-5.0$ in $1,400$s.  In (c), we vary the number of sampling steps per iteration from $5$ to $100$ (we omit the results of Gibbs-1 since it diverges with less than 100 steps) and report the RMSE after 10,000 iterations. DMALA and DULA outperform GWG consistently and the improvement becomes larger when the number of sampling steps becomes smaller, demonstrating the  fast mixing of our discrete Langevin proposal. 

\begin{table*}[h]
    \centering
    \begin{tabular}{c|c|cccccccc}
    \hline \hline
        \multirow{3}{*}{Model} & \multirow{3}{*}{Methods} & \multirow{3}{*}{Self-BLEU ($\downarrow$)} & \multicolumn{6}{c}{Unique $n$-grams ($\%$) ($\uparrow$)} &  \multirow{3}{*}{Corpus BLEU ($\uparrow$)}  \\ 
        \cline{4-9}
        & & & \multicolumn{2}{c}{Self} &\multicolumn{2}{c}{WT103} & \multicolumn{2}{c}{TBC} & \\
        \cline{4-9}
        & & & $n=2$ & $n=3$ & $n=2$ & $n=3$ & $n=2$ & $n=3$ & \\ \hline
        \multirow{3}{*}{Bert-Base} 
        & Gibbs & 86.84 & 10.98 & 16.08 & 18.57 & 32.21 & 21.22 & 33.05 & \textbf{23.82} \\
        & GWG & 81.97 & 15.12 & 21.79 & 22.76 & 37.59 & 24.72 & 37.98 & 22.84   \\
        & DULA & \textbf{72.37} & \textbf{23.33} & \textbf{32.88} & 27.74 & \textbf{45.85} & 30.02 & \textbf{46.75} & 21.82 \\
        & DMALA & 72.59 & 23.26 & 32.64 & \textbf{27.99} & 45.77 & \textbf{30.32} & 46.49 & 21.85 \\ \hline
        \multirow{3}{*}{Bert-Large} 
        & Gibbs & 88.78 & 9.31 & 13.74 & 17.78 & 30.50 & 20.48 & 31.23 & 22.57 \\
        & GWG & 86.50 & 11.03 & 16.13 & 19.25 & 33.20 & 21.42 & 33.54 & \textbf{23.08} \\
        & DULA & 77.96 & 17.97 & 26.64 & 23.69 & 41.30 & 26.18 & 42.14 & 21.28 \\
        & DMALA & \textbf{76.27} & \textbf{19.83} & \textbf{28.48} & \textbf{25.38} & \textbf{42.94} & \textbf{27.87} & \textbf{43.77} & 21.73 \\
    \hline \hline
    \end{tabular}
    \caption{Quantative results on text infilling. The reference text for computing the Corpus BLEU is the combination of WT103 and TBC. DULA and DMALA generate sentences with a similar level of quality as Gibbs and GWG (measured by Corpus BLEU) while attaining higher diversity (measured by self-BLEU and Unique $n$-grams).}
    \label{tab:text}
\end{table*}

\begin{figure*}[h]
    \centering
    \includegraphics[width=0.98\textwidth]{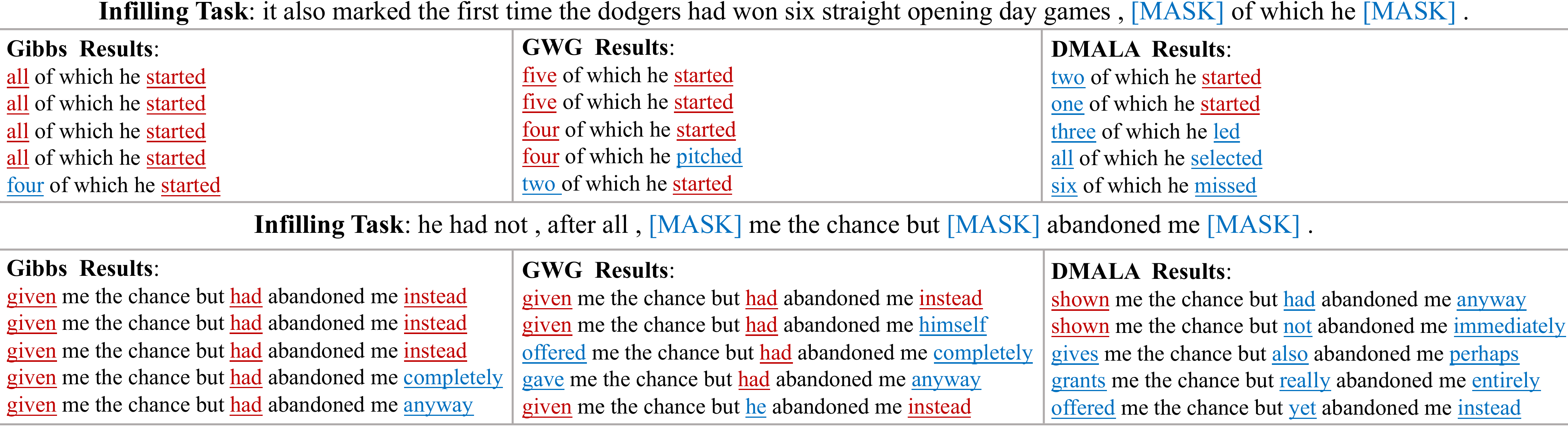}
    \caption{Examples of the generated sentences on text infilling. {\color{cyan} Blue} and \red{Red} words are generated where \red{Red} indicates repetitive generation. DMALA can generate semantically meaningful sentences with much higher diversity.}
    \label{fig:generate_results}
\end{figure*}

\subsubsection{Deep EBMs}

We train deep EBMs where $E_\theta$ is a ResNet~\cite{he2016deep} with Persistent Contrastive Divergence~\cite{tieleman2008training, tieleman2009using} and a replay buffer~\cite{du2019implicit} following~\citet{grathwohl2021oops}. We run DMALA and DULA for 40 steps per iteration. After training, we adopt Annealed Importance Sampling~\cite{neal2001annealed} to estimate the likelihood. The results for GWG and Gibbs are taken from~\citet{grathwohl2021oops}, and for VAE are taken from~\citet{tomczak2018vae}.

\textbf{Results} In Table~\ref{tab:ebm}, we see that DMALA yields the highest log-likelihood among all methods and its generated images in Figure~\ref{fig:sample-ebm} in Appendix~\ref{sec:appendix-learned-ebm} are very close to the true images. DULA runs the same number of steps as DMALA and GWG but only with half of the cost. 
We hypothesis that running DULA for more steps or with an adaptive stepsize schedule~\citep{song2019generative} will improve its performance.

\subsection{Binary Bayesian Neural Networks}\label{sec:bnn}

Bayesian neural networks have been shown to provide strong predictions and uncertainty estimation in deep learning~\cite{hernandez2015probabilistic, zhang2019cyclical, liu2021sampling}. In the meanwhile, binary neural networks~\cite{courbariaux2016binarized,  rastegari2016xnor, liu2021post}, i.e. the weight is in $\{-1, 1\}$, accelerate the learning and significantly reduce computational and memory costs. To combine the benefits of both worlds, we consider training a binary Bayesian neural network with discrete sampling.
We conduct regression on four UCI datasets~\cite{Dua:2019}, and the energy function is defined as,
\begin{equation*}
    U(\theta) = -\sum_{i=1}^N ||f_{\theta}(x_i) - y_i||^2,
\end{equation*}
where $D=\{x_i, y_i\}_{i=1}^N$ is the training dataset, and $f_\theta$ represents a two-layer neural network with \texttt{Tanh} activation and 500 hidden neurons. The dimension of the weight $d$ varies on different datasets ranging from $7,500$ to $45,000$. We report the log-likelihood on the training set together with root mean-square-error (RMSE) on the test set.

\textbf{Results} From Table~\ref{tab:bayesbnn}, we observe that DMALA and DULA outperform other methods significantly on all datasets except test RMSE on COMPAS, which we hypothesize is because of overfitting (the training set only has $4,900$ data points). These results demonstrate that our methods converge fast for high dimensional distributions, due to the ability to make large moves per iteration, and suggest that our methods are compelling for training low-precision Bayesian neural networks of which the weight is discrete.

\subsection{Text Infilling}

Text infilling is an important and intriguing task where the goal is to fill in the blanks given the context~\cite{zhu2019text, donahue2020enabling}. Prior work has realized it by sampling from a categorical distribution produced by BERT~\cite{devlin2018bert,wang2019bert}. However, there are a huge number of word combinations, which makes sampling from the categorical distribution difficult. Therefore, an efficient sampler is needed to producing high-quality text.

We randomly sample 20 sentences from TBC~\cite{zhu2015aligning} and WiKiText-103~\cite{merity2016pointer}, mask 25\% of the words in the sentence~\citep{zhu2019text, donahue2020enabling}, and sample 25 sentences from the probability distribution given by BERT. We run all samplers for 50 steps based on two models, BERT-base and BERT-large. As a common practice in non-autoregressive text generation, we select the top-5 sentences with the highest likelihood out of the 25 sentences to avoid low-quality generation~\cite{gu2018non, zhou2019understanding}. We evaluate the methods from two perspectives, diversity and quality. For diversity, we use self-BLEU~\cite{zhu2018texygen} and the number of unique n-grams~\cite{wang2019bert} to measure the difference between the generated sentences. For quality, we measure the BLEU score~\cite{papineni2002bleu} between the generated texts and the original dataset (TBC+WikiText-103)~\citep{wang2019bert,yu2017seqgan}. Note that the BLEU score can only approximately represent the quality of the generation since it cannot handle out-of-distribution generations. We use one-hot vectors to represent categorical variables. We put the discussion of DLP with categorical variables in practice in Appendix~\ref{sec:one-hot}.

\textbf{Results} The quantitative and qualitative results are shown in Table~\ref{tab:text} and Figure~\ref{fig:generate_results}.
We find that DULA and DMALA can fill in the blanks with similar quality as Gibbs and GWG but with much higher diversity. Due to the nature of languages, there exist strong correlations among words. It is generally difficult to change one word given the others fixed while still fulfilling the context. However, it is likely to have another combination of words, which are all different from the current ones, to satisfy the infilling. Because of this, the ability to update all coordinates in one step makes our methods especially suitable for this task, as reflected in the evaluation metrics and generated sentences.

\section{Conclusion}
We propose a Langevin-like proposal for efficiently sampling from complex high-dimensional discrete distributions. Our method, discrete Langevin proposal (DLP), is able to explore discrete structure effectively based on the gradient information.
For different usage scenarios, we have developed several variants with DLP, including unadjusted, Metropolis-adjusted, stochastic, and preconditioned versions. We prove the asymptotic convergence of DLP without the MH step under log-quadratic and general distributions. Empirical results on many different problems demonstrate the superiority of our method over baselines in general settings.

While the Langevin algorithm has achieved great success in continuous spaces, there has always lacked a counterpart of such simple, effective and general-purpose samplers in discrete spaces. We hope our method sheds light on building practical and accurate samplers for discrete distributions.

\section*{Acknowledgements}
We would like to thank Yingzhen Li and the anonymous reviewers for their thoughtful comments on the manuscript. This research is supported by CAREER-1846421, SenSE2037267, EAGER-2041327, Office of Navy Research, and NSF AI Institute for Foundations of Machine Learning (IFML).
\bibliography{example_paper}
\bibliographystyle{icml2022}

\newpage
\appendix
\onecolumn

\section{Algorithm with Binary Variables}
When the variable domain $\Theta$ is binary $\{0,1\}^d$, we could simplify Algorithm~\ref{alg:dlp} further and obtain Algorithm~\ref{alg:binary}, which clearly shows that our method can be cheaply computed in parallel on CPUs and GPUs.

\begin{algorithm}[H]
  \caption{DULA and DMALA with Binary Variables.}
  \begin{algorithmic}
    \label{alg:binary}
    \STATE \textbf{given:} Stepsize $\alpha$.
    
      \LOOP
      \STATE \textbf{compute} $P(\theta) = \frac{\exp(-\frac{1}{2}\nabla U(\theta)\odot (2\theta-1) - \frac{1}{2\alpha})}{\exp(-\frac{1}{2}\nabla U(\theta)\odot (2\theta-1) - \frac{1}{2\alpha})+1}$
      \STATE \textbf{sample} $\mu \sim \text{Unif}(0,1)^d$
      \STATE $I \leftarrow \texttt{dim}(\mu\le P(\theta))$
    \STATE $\theta'\leftarrow \texttt{flipdim}(I)$
    
    \vspace{0.5em}
    \STATE $\triangleright$ Optionally, do the MH step
    \STATE \textbf{compute} $q(\theta'|\theta) = \prod_i q_i(\theta'_i|\theta)=\prod_{i\in I} P(\theta)_i \cdot \prod_{i\notin I} (1-P(\theta)_i)$
    \STATE \textbf{compute} $P(\theta') = \frac{\exp(-\frac{1}{2}\nabla U(\theta')\odot (2\theta'-1) - \frac{1}{2\alpha})}{\exp(-\frac{1}{2}\nabla U(\theta')\odot (2\theta'-1) - \frac{1}{2\alpha})+1}$
    \STATE \textbf{compute} $q(\theta|\theta') = \prod_i q_i(\theta_i|\theta')=\prod_{i\in I} P(\theta')_i \cdot \prod_{i\notin I} (1-P(\theta')_i)$
    \STATE \textbf{set} $\theta \leftarrow \theta'$ with probability
    \[
    \min\left(1,\exp\left(U(\theta')-U(\theta)\right) \frac{q(\theta|\theta')}{q(\theta'|\theta)}\right)
    \]
    \ENDLOOP
    \STATE \textbf{output}: samples $\{\theta_k\}$
    
  \end{algorithmic}
\end{algorithm}

\section{Algorithm with Categorical Variables}\label{sec:one-hot}
When using one-hot vectors to represent categorical variables, our discrete Langevin proposal becomes

\begin{align*}
   \text{Categorical}\Big(\text{Softmax}\Big(\frac{1}{2}\nabla U(\theta)_i^{\intercal}(\theta_i'-\theta_i)
    -\frac{\norm{\theta_i'-\theta_i}_2^2}{2\alpha}\Big)\Big),
\end{align*}
where $\theta_i,\theta'_i$ are one-hot vectors.

However, if the variables are ordinal with clear ordering information, we can also use integer representation $\theta\in\{0, 1, \ldots, S-1\}^d$ and compute DLP as in Equation\eqref{eq:proposal2}.

\section{Extension to Infinite Domains}\label{sec:infinite}
We only consider finite discrete distributions in the paper. However, our algorithms can be easily extended to infinite distributions. One way to do so is to add a window size for the proposal such that the proposal only considers a local region in each step. For example, if the domain is infinite integer-valued: $\Theta = \{0,1,2,\ldots\}^d$, then our DLP can use a proposal window of $\theta'_i\in\{\theta_i-1,\theta_i,\theta_i+1\}$. It is also possible to use other window sizes. We leave the comprehensive study of our method on infinite discrete distributions for future work. 

\section{Importance of the Term $\norm{\theta'-\theta}^2/(2\alpha)$}\label{sec:stepsize-term}
Similar to DLP, we can also use a first-order Taylor series approximation to Equation~\eqref{eq:local-balanced} and get the following proposal 
\begin{align*}
   \text{Categorical}\Big(\text{Softmax}\Big(\frac{1}{2}\nabla U(\theta)_i(\theta_i'-\theta_i)
    \Big)\Big).
\end{align*}
The above proposal does not contain the stepsize term as in DLP, which will result in a very low acceptance probability in practice. For example, in the RBM experiment (Section~\ref{sec:exp:rbm}), its acceptance probability is $0.005\%$ while DMALA's acceptance probability is $51\%$ on average. This demonstrates the importance of the term $\norm{\theta'-\theta}^2/(2\alpha)$ in DLP where the stepsize $\alpha$ can control the closeness of $\theta'$ to the current $\theta$. The effect of $\alpha$ is similar to the stepsize in the standard Langevin algorithm, and by tuning it our method can achieve a desirable acceptance probability leading to efficient sampling. 

\section{Proof of Theorem~\ref{thm:log-quadratic}}

\begin{proof}
We divide the proof into two parts. In the fisrt part, we will prove the weak convergence and in the second part we will prove the convergence rate with respect to the stepsize $\alpha$.
\paragraph{Weak Convergence.}
The main idea of the proof is to replace the gradient term in the proposal by the energy difference $U(\theta')-U(\theta)$ using Taylor series approximation, and then show the reversibility of the chain based on the proof of Theorem 1 in \citet{zanella2020informed} .

Recall that the target distribution is $\pi(\theta)= \exp\left(\theta^{\intercal} W\theta + b^{\intercal}\theta\right)/Z$. We have that $\nabla U(\theta) = 2W^{\intercal}\theta + b$, $\nabla^2U(\theta)=2W$. Since $\nabla^2 U$ is a constant, we can rewrite the proposal distribution as the following
\begin{align*}
    q_{\alpha}(\theta'|\theta)&=\frac{\exp\left(\frac{1}{2} \nabla U(\theta)^{\intercal}(\theta'-\theta) - \frac{1}{2\alpha}\norm{\theta'-\theta}^2\right)}{\sum_x \exp\left(\frac{1}{2} \nabla U(\theta)^{\intercal}(x-\theta) - \frac{1}{2\alpha}\norm{x-\theta}^2\right)}\\
    &=
    \frac{\exp\left(\frac{1}{2} \nabla U(\theta)^{\intercal}(\theta'-\theta) + \frac{1}{2}(\theta'-\theta)^{\intercal}W (\theta'-\theta) -(\theta'-\theta)^{\intercal}(\frac{1}{2\alpha}I+ \frac{1}{2}W) (\theta'-\theta)\right)}{\sum_x \exp\left(\frac{1}{2} \nabla U(\theta)^{\intercal}(x-\theta) + \frac{1}{2}(x-\theta)^{\intercal}W (x-\theta) -(x-\theta)^{\intercal}(\frac{1}{2\alpha}I+ \frac{1}{2}W)(x-\theta)\right)}\\
    &=
    \frac{\exp\left(\frac{1}{2} \left(U(\theta')-U(\theta)\right) -(\theta'-\theta)^{\intercal}(\frac{1}{2\alpha}I+ \frac{1}{2}W) (\theta'-\theta)\right)}{\sum_x \exp\left(\frac{1}{2} \left(U(x)-U(\theta)\right) -(x-\theta)^{\intercal}(\frac{1}{2\alpha}I+ \frac{1}{2}W)(x-\theta)\right)}
\end{align*}
where the last equation is because $U(\theta')-U(\theta)=\nabla U(\theta)^{\intercal}(\theta'-\theta) + \frac{1}{2}(\theta'-\theta)^{\intercal}2W (\theta'-\theta)$  by Taylor series approximation.

Let $Z_{\alpha}(\theta)=\sum_x \exp\left(\frac{1}{2} \left(U(x)-U(\theta)\right) -(x-\theta)^{\intercal}(\frac{1}{2\alpha}I+ \frac{1}{2}W)(x-\theta)\right)$, and $\pi_{\alpha} = \frac{Z_{\alpha}(\theta)\pi(\theta)}{\sum_x Z_{\alpha}(x)\pi(x)}$, now we will show that $q_{\alpha}$ is reversible w.r.t. $\pi_{\alpha}$.

We have that 
\begin{align*}
    \pi_{\alpha}(\theta)q_{\alpha}(\theta'|\theta)&=\frac{Z_{\alpha}(\theta)\pi(\theta)}{\sum_x Z_{\alpha}(x)\pi(x)}\cdot \frac{\exp\left(\frac{1}{2} \left(U(\theta')-U(\theta)\right) -(\theta'-\theta)^{\intercal}(\frac{1}{2\alpha}I+ \frac{1}{2}W) (\theta'-\theta)\right)}{Z_{\alpha}(\theta)}\\
    &=
    \frac{\exp\left(\frac{1}{2} \left(U(\theta')+U(\theta)\right) -(\theta'-\theta)^{\intercal}(\frac{1}{2\alpha}I+ \frac{1}{2}W) (\theta'-\theta)\right)}{Z\cdot\sum_x Z_{\alpha}(x)\pi(x)}.
\end{align*}

It is clear that this expression is symmetric in $\theta$ and $\theta'$. Therefore $q_{\alpha}$ is reversible and the stationary distribution is $\pi_{\alpha}$.

Now we will prove that $\pi_{\alpha}$ converges weakly to $\pi$ as $\alpha\rightarrow 0$. Notice that for any $\theta$,
\begin{align*}
    Z_{\alpha}(\theta)&=\sum_x \exp\left(\frac{1}{2} \left(U(x)-U(\theta)\right) -(x-\theta)^{\intercal}\left(\frac{1}{2\alpha}I+ \frac{1}{2}W\right)(x-\theta)\right)\\
    &\xRightarrow{\alpha\downarrow 0} 
    \sum_x \exp\left(\frac{1}{2} \left(U(x)-U(\theta)\right)\right)\delta_{\theta}(x)\\
    &=
    1,
\end{align*}
where $\delta_{\theta}(x)$ is a Dirac delta. It follows that $\pi_{\alpha}$ converges pointwisely to $\pi(\theta)$. By Scheffé’s Lemma, we attain that $\pi_{\alpha}$ converges weakly to $\pi$.

\paragraph{Convergence Rate w.r.t. Stepsize.}
Let us consider the convergence rate in terms of the $L_1$-norm 

\begin{align*}
    \norm{\pi_{\alpha} -\pi}_1= \sum_{\theta}\Abs{\frac{Z_{\alpha}(\theta)\pi(\theta)}{\sum_x Z_{\alpha}(x)\pi(x)} - \pi(\theta)}.
\end{align*}

We write out each absolute value term 
\begingroup\makeatletter\def\f@size{8}\check@mathfonts
\def\maketag@@@#1{\hbox{\m@th\large\normalfont#1}}
\begin{align*}
     \Abs{\frac{Z_{\alpha}(\theta)\pi(\theta)}{\sum_x Z_{\alpha}(x)\pi(x)}-\pi(\theta)
     }
     &=
      \pi(\theta)\Abs{\frac{Z_{\alpha}(\theta)}{\sum_x Z_{\alpha}(x)\pi(x)}-1
     }\\
     &=
     \pi(\theta)\Abs{
     \frac{1+\sum_{x\neq \theta}\exp\left(\frac{1}{2} U(x)-\frac{1}{2} U(\theta)-(x-\theta)^{\intercal}\left(\frac{1}{2\alpha}I+\frac{1}{2}W\right)(x-\theta)\right)}
     {1+\sum_y \frac{1}{Z}\exp\left(U(y)\right)\sum_{x\neq y}\exp\left(\frac{1}{2} U(x)-\frac{1}{2} U(y)-(x-y)^{\intercal}\left(\frac{1}{2\alpha}I+\frac{1}{2}W\right)(x-y)\right)} -1
     }.
\end{align*}
\endgroup

Since $\lambda_{\text{min}}(W)\norm{x}^2\le x^{\intercal}Wx, \forall x$, it follows that
\[
(x-\theta)^{\intercal}\left(\frac{1}{2\alpha}I+\frac{1}{2}W\right)(x-\theta)\ge \frac{1+\alpha\lambda_{\text{min}}}{2\alpha}\norm{x-\theta}^2.
\]

We also notice that $\min_{x\neq \theta}\norm{x-\theta}^2 = 1$, thus when $\frac{Z_{\alpha}(\theta)}{\sum_x Z_{\alpha}(x)\pi(x)}-1>0$, we get
\begingroup\makeatletter\def\f@size{8}\check@mathfonts
\def\maketag@@@#1{\hbox{\m@th\large\normalfont#1}}
\begin{align*}
     \Abs{\frac{Z_{\alpha}(\theta)\pi(\theta)}{\sum_x Z_{\alpha}(x)\pi(x)}-\pi(\theta)
     }
     &=
     \pi(\theta)\left(
     \frac{1+\sum_{x\neq \theta}\exp\left(\frac{1}{2} U(x)-\frac{1}{2} U(\theta)-(x-\theta)^{\intercal}\left(\frac{1}{2\alpha}I+\frac{1}{2}W\right)(x-\theta)\right)}
     {1+\sum_y \frac{1}{Z}\exp\left(U(y)\right)\sum_{x\neq y}\exp\left(\frac{1}{2} U(x)-\frac{1}{2} U(y)-(x-y)^{\intercal}\left(\frac{1}{2\alpha}I+\frac{1}{2}W\right)(x-y)\right)} -1
     \right)\\
     &\le
     \pi(\theta)\left(
     1+\sum_{x\neq \theta}\exp\left(\frac{1}{2} U(x)-\frac{1}{2} U(\theta)-\frac{1+\alpha\lambda_{\text{min}}}{2\alpha}\norm{x-\theta}^2\right)
      -1
     \right)\\
     &\le
     \pi(\theta)\left(
     1+\exp\left(-\frac{1+\alpha\lambda_{\text{min}}}{2\alpha}\right)\sum_{x\neq \theta}\exp\left(\frac{1}{2} U(x)-\frac{1}{2} U(\theta)\right)
      -1
     \right)\\
     &=
     \pi(\theta)\left(
     \sum_{x\neq \theta}\exp\left(\frac{1}{2} U(x)-\frac{1}{2} U(\theta)\right)
     \right)\cdot\exp\left(-\frac{1+\alpha\lambda_{\text{min}}}{2\alpha}\right)\\
     &\le
     \pi(\theta)\left(
     \sum_{x}\exp\left(U(x)\right)
     \right)\cdot\exp\left(-\frac{1+\alpha\lambda_{\text{min}}}{2\alpha}\right)\\
     &=
     \pi(\theta)Z
     \cdot\exp\left(-\frac{1+\alpha\lambda_{\text{min}}}{2\alpha}\right).
\end{align*}
\endgroup

Similarly, when $\frac{Z_{\alpha}(\theta)}{\sum_x Z_{\alpha}(x)\pi(x)}-1<0$, we have,
\begingroup\makeatletter\def\f@size{8}\check@mathfonts
\def\maketag@@@#1{\hbox{\m@th\large\normalfont#1}}
\begin{align*}
     \Abs{\frac{Z_{\alpha}(\theta)\pi(\theta)}{\sum_x Z_{\alpha}(x)\pi(x)}-\pi(\theta)
     }
     &=
     \pi(\theta)\left(1-
     \frac{1+\sum_{x\neq \theta}\exp\left(\frac{1}{2} U(x)-\frac{1}{2} U(\theta)-(x-\theta)^{\intercal}\left(\frac{1}{2\alpha}I+\frac{1}{2}W\right)(x-\theta)\right)}
     {1+\sum_y \frac{1}{Z}\exp\left(U(y)\right)\sum_{x\neq y}\exp\left(\frac{1}{2} U(x)-\frac{1}{2} U(y)-(x-y)^{\intercal}\left(\frac{1}{2\alpha}I+\frac{1}{2}W\right)(x-y)\right)} 
     \right)\\
     &\le
    \pi(\theta)\left(1 -
     \frac{1}
     {1+\sum_y \frac{1}{Z}\exp\left(U(y)\right)\sum_{x\neq y}\exp\left(\frac{1}{2} U(x)-\frac{1}{2} U(y)-\frac{1+\alpha\lambda_{\text{min}}}{2\alpha}\right)}
     \right)\\
     &=
     \pi(\theta)\left(
     \frac{\sum_y \frac{1}{Z}\exp\left(U(y)\right)\sum_{x\neq y}\exp\left(\frac{1}{2} U(x)-\frac{1}{2} U(y)-\frac{1+\alpha\lambda_{\text{min}}}{2\alpha}\right)}
     {1+\sum_y \frac{1}{Z}\exp\left(U(y)\right)\sum_{x\neq y}\exp\left(\frac{1}{2} U(x)-\frac{1}{2} U(y)-\frac{1+\alpha\lambda_{\text{min}}}{2\alpha}\right)}
     \right)\\
     &\le
     \pi(\theta)\left(
     \sum_y \frac{1}{Z}\exp\left(U(y)\right)\sum_{x\neq y}\exp\left(\frac{1}{2} U(x)-\frac{1}{2} U(y)\right)
     \right)\cdot\exp\left(-\frac{1+\alpha\lambda_{\text{min}}}{2\alpha}\right)\\
     &\le
     \pi(\theta)\left(
     \sum_{x}\exp\left(U(x)\right)
     \right)\cdot\exp\left(-\frac{1+\alpha\lambda_{\text{min}}}{2\alpha}\right)\\
     &=
     \pi(\theta)Z
     \cdot\exp\left(-\frac{1+\alpha\lambda_{\text{min}}}{2\alpha}\right).
\end{align*}
\endgroup

Therefore, the difference between $\pi_{\alpha}$ and $\pi$ can be bounded as follows
\begin{align*}
    \norm{\pi_{\alpha} -\pi}_1\le \sum_{\theta}\pi(\theta)Z
     \cdot\exp\left(-\frac{1+\alpha\lambda_{\text{min}}}{2\alpha}\right)=Z
     \cdot\exp\left(-\frac{1+\alpha\lambda_{\text{min}}}{2\alpha}\right).
\end{align*}

\end{proof}

\section{Proof of Theorem~\ref{thm:general}}

\begin{proof}
We use a log-quadratic distribution that is close to $\pi$ as an intermediate term to bound the bias of DULA. Recall that $\pi$ is the target distribution, $\pi'$ is the log-quadratic distribution that is close to $\pi$ and $\pi_{\alpha}$ is the stationary distribution of DULA. We let $\pi'_{\alpha}$ be the stationary distributions of DULA targeting $\pi'$, then by triangle inequality,
\begin{align*}
    \norm{\pi_{\alpha} -\pi}_1
    &\le 
    \norm{\pi_{\alpha}-\pi'_{\alpha}}_1 + \norm{\pi'_{\alpha}-\pi'}_1 + \norm{\pi'-\pi}_1.
\end{align*}

\paragraph{Bounding $\norm{\pi'-\pi}_1$.} Let the energy function of $\pi'$ be $V(\theta) = \theta^{\intercal}W\theta + b\theta$.
Since $\Theta$ is a discrete space, there exists a bounded subset $\Omega\in\RR^d$ such that $\Theta$ is a subset of $\Omega$. By Poincaré inequality, we get
\[
\Abs{U(\theta)-V(\theta)}\le C_1 \cdot\norm{\nabla U(\theta) - (W\theta+b)}_1\le C_1\epsilon, \hspace{2em} \forall \theta\in\Theta
\]
where the constant $C_1$ depends on $\Omega$.

Recall that $\pi(\theta) = \frac{\exp(U(\theta))}{Z}$. Let $\pi'(\theta) = \frac{\exp(V(\theta))}{Z'}$ where $Z'$ is the normalizing constant to make $\pi'$ a distribution. Then
\begin{align}
    \norm{\pi-\pi'}_1 
    &=
    \sum_{\theta\in \Theta}\Abs{\frac{\exp(U(\theta))}{Z} - \frac{\exp(V(\theta))}{Z'}}\label{eq:norm1}.
\end{align}
We notice that $\forall \theta$,
\[
\frac{\exp(U(\theta))}{\exp(V(\theta))}=\exp(U(\theta)-V(\theta))\le \exp(\Abs{U(\theta)-V(\theta)})\le\exp(C\epsilon),
\]
and similarly
\[
\frac{\exp(V(\theta))}{\exp(U(\theta))}\le\exp(C\epsilon).
\]
Therefore we have $\forall \theta$,
\begin{align}
    \Abs{\frac{\exp(U(\theta))}{Z} - \frac{\exp(V(\theta))}{Z'}} 
    &=
    \Abs{\frac{\exp(U(\theta))\cdot N - \exp(V(\theta))\cdot Z}{Z\cdot N}}\nonumber\\ 
    &\le 
    \left(\exp(2C\epsilon)-1\right) \max\left\{\frac{\exp(U(\theta))}{Z}, \frac{\exp(V(\theta))}{Z'}\right\}\nonumber\\
    &\le 
    \left(\exp(2C\epsilon)-1\right) \left(\frac{\exp(U(\theta))}{Z}+ \frac{\exp(V(\theta))}{Z'}\right)\label{eq:ineq}.
\end{align}
Plugging the above in Equation~\eqref{eq:norm1}, we obtain 
\begin{align*}
    \norm{\pi-\pi'}_1 &\le
    2\left(\exp(2C\epsilon)-1\right).
\end{align*}

\paragraph{Bounding $\norm{\pi'_{\alpha}-\pi'}_1$.} By Theorem~\ref{thm:log-quadratic}, we know
\begin{align*}
    \norm{\pi'_{\alpha}-\pi'}_1 &\le
    Z'\cdot\exp\left(-\frac{1+\alpha\lambda_{\text{min}}}{2\alpha}\right).
\end{align*}

\paragraph{Bounding $\norm{\pi_{\alpha}-\pi'_{\alpha}}_1$.}
Now we will bound $\norm{\pi_{\alpha}-\pi'_{\alpha}}_1$ by perturbation bounds. Let $T_{\alpha}$, $\tilde{T}_{\alpha}$  be the transition matrices of DULA on $\pi$ and $\pi'$ respectively. We consider the difference between these two matrices.
\begin{align*}
    \norm{T_{\alpha}-\tilde{T}_{\alpha}}_{\infty} = \max_{\theta} \sum_{\theta'} \Abs{
    \frac{\exp\left(\frac{1}{2}\nabla U(\theta)^{\intercal}(\theta'-\theta)-\frac{1}{2\alpha}\norm{\theta'-\theta}^2\right)}{Z_{\alpha}(\theta)}
    -
    \frac{\exp\left(\frac{1}{2}\nabla V(\theta)^{\intercal}(\theta'-\theta)-\frac{1}{2\alpha}\norm{\theta'-\theta}^2\right)}{Z'_{\alpha}(\theta)}}
\end{align*}
where
\[
Z_{\alpha}(\theta) = \sum_{x}\exp\left(\frac{1}{2}\nabla U(\theta)^{\intercal}(x-\theta)-\frac{1}{2\alpha}\norm{x-\theta}^2\right),
Z'_{\alpha}(\theta) = \sum_{x}\exp\left(\frac{1}{2}\nabla V(\theta)^{\intercal}(x-\theta)-\frac{1}{2\alpha}\norm{x-\theta}^2\right).
\] 
We denote $D=\max_{\theta,\theta'\in \Theta}\norm{\theta'-\theta}_{\infty}$. By the assumption $\norm{\nabla U(\theta) - \nabla V(\theta)}_1\le \epsilon$, similar to Equation~\eqref{eq:ineq}, we have
\begin{align*}
    &\Abs{
    \frac{\exp\left(\frac{1}{2}\nabla U(\theta)^{\intercal}(\theta'-\theta)-\frac{1}{2\alpha}\norm{\theta'-\theta}^2\right)}{Z_{\alpha}(\theta)}
    -
    \frac{\exp\left(\frac{1}{2}\nabla V(\theta)^{\intercal}(\theta'-\theta)-\frac{1}{2\alpha}\norm{\theta'-\theta}^2\right)}{Z'_{\alpha}(\theta)}}\\
    &\le
    \left(\exp\left(D\epsilon\right)-1\right)\\
    &\hspace{2em}\cdot\max\left\{\frac{\exp\left(\frac{1}{2}\nabla U(\theta)^{\intercal}(\theta'-\theta)-\frac{1}{2\alpha}\norm{\theta'-\theta}^2\right)}{Z_{\alpha}(\theta)}
    ,
    \frac{\exp\left(\frac{1}{2}\nabla V(\theta)^{\intercal}(\theta'-\theta)-\frac{1}{2\alpha}\norm{\theta'-\theta}^2\right)}{Z'_{\alpha}(\theta)}\right\}\\
    &\le
    \left(\exp\left(D\epsilon\right)-1\right)\\
    &\hspace{2em}\cdot\left(\frac{\exp\left(\frac{1}{2}\nabla U(\theta)^{\intercal}(\theta'-\theta)-\frac{1}{2\alpha}\norm{\theta'-\theta}^2\right)}{Z_{\alpha}(\theta)}
    +
    \frac{\exp\left(\frac{1}{2}\nabla V(\theta)^{\intercal}(\theta'-\theta)-\frac{1}{2\alpha}\norm{\theta'-\theta}^2\right)}{Z'_{\alpha}(\theta)}\right).
\end{align*}

Now we substitute it to $\norm{T_{\alpha}-\tilde{T}_{\alpha}}_{\infty}$,
\begin{align*}
    \norm{T-\tilde{T}}_{\infty} &\le 
    2\left(\exp\left(D\epsilon\right)-1\right).
\end{align*}

By the perturbation bound in \citet{schweitzer1968perturbation},
\[
\norm{\pi_{\alpha}-\pi'_{\alpha}}_1\le C_2\cdot\norm{T_{\alpha}-\tilde{T}_{\alpha}}_{\infty}=2C_2\left(\exp\left(D\epsilon\right)-1\right)
\]
where $C_2$ is a constant depending on $\pi'$ and $\alpha$. Please note that it is also possible to use other perturbation bounds~\citep{cho2001comparison}.

Combining these three bounds, we get 
\begin{align*}
    \norm{\pi_{\alpha} -\pi}_1
    &\le 
    \norm{\pi_{\alpha}-\pi'_{\alpha}}_1 + \norm{\pi'_{\alpha}-\pi'}_1 + \norm{\pi'-\pi}_1\\
    &\le 
    2C_2\left(\exp\left(D\epsilon\right)-1\right) + Z'\cdot\exp\left(-\frac{1+\alpha\lambda_{\text{min}}}{2\alpha}\right) + 2\left(\exp(2C_1\epsilon)-1\right).
\end{align*}
We define $c_1:= 2\max (2, 2C_2)$ and $c_2 := \max(2C_1, D)$, then we reach the the final result
\begin{align*}
    \norm{\pi_{\alpha} -\pi}_1
    &\le 
    \norm{\pi_{\alpha}-\pi'_{\alpha}}_1 + \norm{\pi'_{\alpha}-\pi'}_1 + \norm{\pi'-\pi}_1\\
    &\le 
    2c_1\cdot\left(\exp\left(c_2\epsilon\right)-1\right) + Z'\cdot\exp\left(-\frac{1+\alpha\lambda_{\text{min}}}{2\alpha}\right).
\end{align*}
\end{proof}

\section{Proof of Theorem~\ref{thm:stochastic}}
\begin{proof}
The proposal distribution for the coordinate $i$ with the stochastic gradient is
\[
\hat{q}_i(\theta'_i|\theta)=\text{Categorical}\left(\text{Softmax}\left(\exp\left(\hat{\nabla}U(\theta)_i(\theta'_i-\theta_i)-\frac{1}{2\alpha}(\theta'_i-\theta_i)^2\right)\right)\right).
\]

We consider the binary case i.e. $\Theta_i=\{0,1\}$. When the current position $\theta_i=0$ the probability for $\theta'_i$ is
\[
\frac{1}{\exp\left(\hat{\nabla}U(\theta)_i-\frac{1}{2\alpha}\right)1}, \hspace{1em}\text{when }\theta'_i=0;
\hspace{2em}
\frac{1}{\exp\left(-\hat{\nabla}U(\theta)_i+\frac{1}{2\alpha}\right)+1}, \hspace{1em}\text{when }\theta'_i=1.
\]

The difference between $\PExv{\hat{q}_i}$ and $q_i$ is
\begin{align*}
    \norm{\PExv{\hat{q}_i} - q_i}_1
    &=
    \sum_{\theta'_i=0}^1\Abs{\PExv{\hat{q}_i(\theta'_i|\theta)} - q_i(\theta'_i|\theta)}.
\end{align*}
We consider each absolute value term. When $\theta'_i=0$, 
\begin{align*}
    \Abs{\PExv{\hat{q}_i(\theta'_i=0|\theta)} - q_i(\theta'_i=0|\theta)}&=
    \Abs{
    \PExv{\frac{1}{\exp\left(\hat{\nabla}U(\theta)_i-\frac{1}{2\alpha}\right)+1}}
    -
    \frac{1}{\exp\left(\nabla U(\theta)_i-\frac{1}{2\alpha}\right)+1}
    }\\
    &\le
    \PExv{\Abs{
    \frac{1}{\exp\left(\hat{\nabla}U(\theta)_i-\frac{1}{2\alpha}\right)+1}
    -
    \frac{1}{\exp\left(\nabla U(\theta)_i-\frac{1}{2\alpha}\right)+1}
    }}.
\end{align*}

Since the absolute value of the derivative for $f(x) = \frac{1}{\exp(x)+1}$ is
$\frac{\exp(x)}{(\exp(x)+1)^2}$,
which is monotonically increasing for $x\in(-\infty,0]$ and monotonically decreasing for $x\in(0,\infty)$. 
Then by the assumption $\Abs{\hat{\nabla}U(\theta)_i}\le L$, $\Abs{\nabla U(\theta)_i}\le L$ and Lipschitz continuity,
\begin{align*}
    \Abs{\PExv{\hat{q}_i(\theta'_i=0|\theta)} - q_i(\theta'_i=0|\theta)}
    &\le
\max\left\{\frac{\exp(-\frac{1}{2\alpha}+L)}{\left(\exp(-\frac{1}{2\alpha}+L)+1\right)^2},\frac{\exp(-\frac{1}{2\alpha}-L)}{\left(\exp(-\frac{1}{2\alpha}-L)+1\right)^2}\right\}\PExv{\Abs{\hat{\nabla}U(\theta)_i - \nabla U(\theta)_i}}\\
    &\le
    \exp\left(-\frac{1}{2\alpha}+L\right)\PExv{\Abs{\hat{\nabla}U(\theta)_i - \nabla U(\theta)_i}}\\
    &\le
    \exp\left(-\frac{1}{2\alpha}+L\right)\sigma
\end{align*}
where the last equation is because of our assumption on the variance of the stochastic gradient $\PVar{\hat{\nabla}U(\theta)_i}\le \sigma^2$, which leads to
\[
\PExv{\Abs{\hat{\nabla}U(\theta)_i - \nabla U(\theta)_i}}^2\le \PExv{\Abs{\hat{\nabla}U(\theta)_i - \nabla U(\theta)_i}^2}=\PVar{\hat{\nabla}U(\theta)_i}\le\sigma^2.
\]
Similarly, when $\theta_i=0$ 
\begin{align*}
    \Abs{\PExv{\hat{q}_i(\theta'_i=1|\theta)} - q_i(\theta'_i=1|\theta)}
    &\le
    \max\left\{\frac{\exp(\frac{1}{2\alpha}-L)}{\left(\exp(\frac{1}{2\alpha}-L)+1\right)^2},\frac{\exp(\frac{1}{2\alpha}+L)}{\left(\exp(\frac{1}{2\alpha}+L)+1\right)^2}\right\}
   \PExv{\Abs{\hat{\nabla}U(\theta)_i - \nabla U(\theta)_i}}\\
    &\le
    \exp\left(-\frac{1}{2\alpha}+L\right)\PExv{\Abs{\hat{\nabla}U(\theta)_i - \nabla U(\theta)_i}}\\
    &\le
    \exp\left(-\frac{1}{2\alpha}+L\right)\sigma.
\end{align*}
We substitute the above two bounds to the $L_1$ distance
\begin{align*}
    \norm{\PExv{\hat{q}_i} - q_i}_1
    &\le 
    2\sigma\cdot\exp\left(-\frac{1}{2\alpha}+L\right).
\end{align*}
The same analysis applies to the case when the current position $\theta_i = 1$. To see this, we write out the proposal distribution 
\[
\frac{1}{\exp\left(\hat{\nabla}U(\theta)_i+\frac{1}{2\alpha}\right)+1}, \hspace{1em}\text{when }\theta'_i=0;
\hspace{2em}
\frac{1}{\exp\left(-\hat{\nabla}U(\theta)_i-\frac{1}{2\alpha}\right)+1}, \hspace{1em}\text{when }\theta'_i=1.
\]
The remaining follows the same as in $\theta_i=0$.

\end{proof}

\section{Preconditioned Discrete Langevin Proposal}\label{sec:preconditioner}
Our discrete Langevin proposal can be easily combined with diagonal preconditioners as in the standard Langevin proposal. We consider the case when the preconditioner is constant.

Assume that the constant diagonal preconditioner is $G=\text{diag}(g)$ where $g\in\RR^d$, then we can derive the preconditioned discrete Langevin proposal by a simple coordinate transformation. Let the new coordinates be $\eta_i := g_i^{-1}\theta_i$, then the domain for $\eta_i$ is $H_i = \{g_i^{-1}\theta_i: \theta_i\in\Theta_i\}$. The discrete Langevin proposal for $\eta_i$ is
\begin{align*}
   q(\eta_i'|\eta)&=
\frac{\exp\left(\frac{g_i}{2}\nabla U(G\eta)_i(\eta_i'-\eta_i)-\frac{(\eta_i'-\eta_i)^2}{2\alpha}\right)}{\sum_{x_i\in H_i}\exp\left(\frac{g_i}{2}\nabla U(G\eta)_i(x_i-\eta_i)-\frac{(x_i-\eta_i)^2}{2\alpha}\right)}.
\end{align*}
Let $\eta'_i=g_i^{-1}\theta'_i, x_i = g_i^{-1}y_i$, then the above proposal distribution is equivalent to
\begin{align*}
   q(\theta'_i|\theta)
&=
\frac{\exp\left(\frac{g_i}{2}\nabla U(\theta)_i(g_i^{-1}\theta'_i-g_i^{-1}\theta_i)-\frac{(g_i^{-1}\theta'_i-g_i^{-1}\theta_i)^2}{2\alpha}\right)}{\sum_{y_i\in\Theta_i}\exp\left(\frac{g_i}{2}\nabla U(\theta)_i(g_i^{-1}y_i-g_i^{-1}\theta_i)-\frac{(g_i^{-1}y_i-g_i^{-1}\theta_i)^2}{2\alpha}\right)}\\
&=
\frac{\exp\left(\frac{1}{2}\nabla U(\theta)_i(\theta'_i-\theta_i)-\frac{(\theta'_i-\theta_i)^2}{2\alpha g_i^2}\right)}{\sum_{y_i\in\Theta_i}\exp\left(\frac{1}{2}\nabla U(\theta)_i(y_i-\theta_i)-\frac{(y_i-\theta_i)^2}{2\alpha g_i^2}\right)}.
\end{align*}
\paragraph{Example.} To demonstrate the use of the preconditioned discrete Langevin proposal, we consider an Ising model $U(\theta) = \theta^{\intercal}W\theta$ where $W=\text{diag}(-0.001, -1000)$ and $\theta\in\{-1,1\}$. Due to the drastically different scale for the derivative of different coordinates, if using the same stepsize for both coordinates, no values will work. However if we use the preconditioned proposal as shown above and let $\alpha=1,g_1=1000, g_2=0.001$. Then the Markov chain will mix quickly. For example, in terms of the log RMSE between the estimated mean and the true mean, preconditioned DLP gives $-6.2$ whereas the standard DLP with any $\alpha$ can only achieve around $-0.34$.

\section{Additional Experiments Results and Setting Details}
\subsection{Ising Model for Theory Verification}
To verify Theorem~\ref{thm:log-quadratic}, we use a 2 by 2 Ising model $U(\theta) = \theta^{\intercal}cW\theta + b\theta$ where $W$ is the binary adjacency matrix and $c=0.1,b=0.2$. To verify Theorem~\ref{thm:general}, we set $a=1$ and $b=0.1$.

\subsection{Sampling from Ising Models}\label{sec:appendix-ising}
\begin{figure}[h!]
\centering
	\begin{tabular}{ccc}		
		\includegraphics[width=6cm]{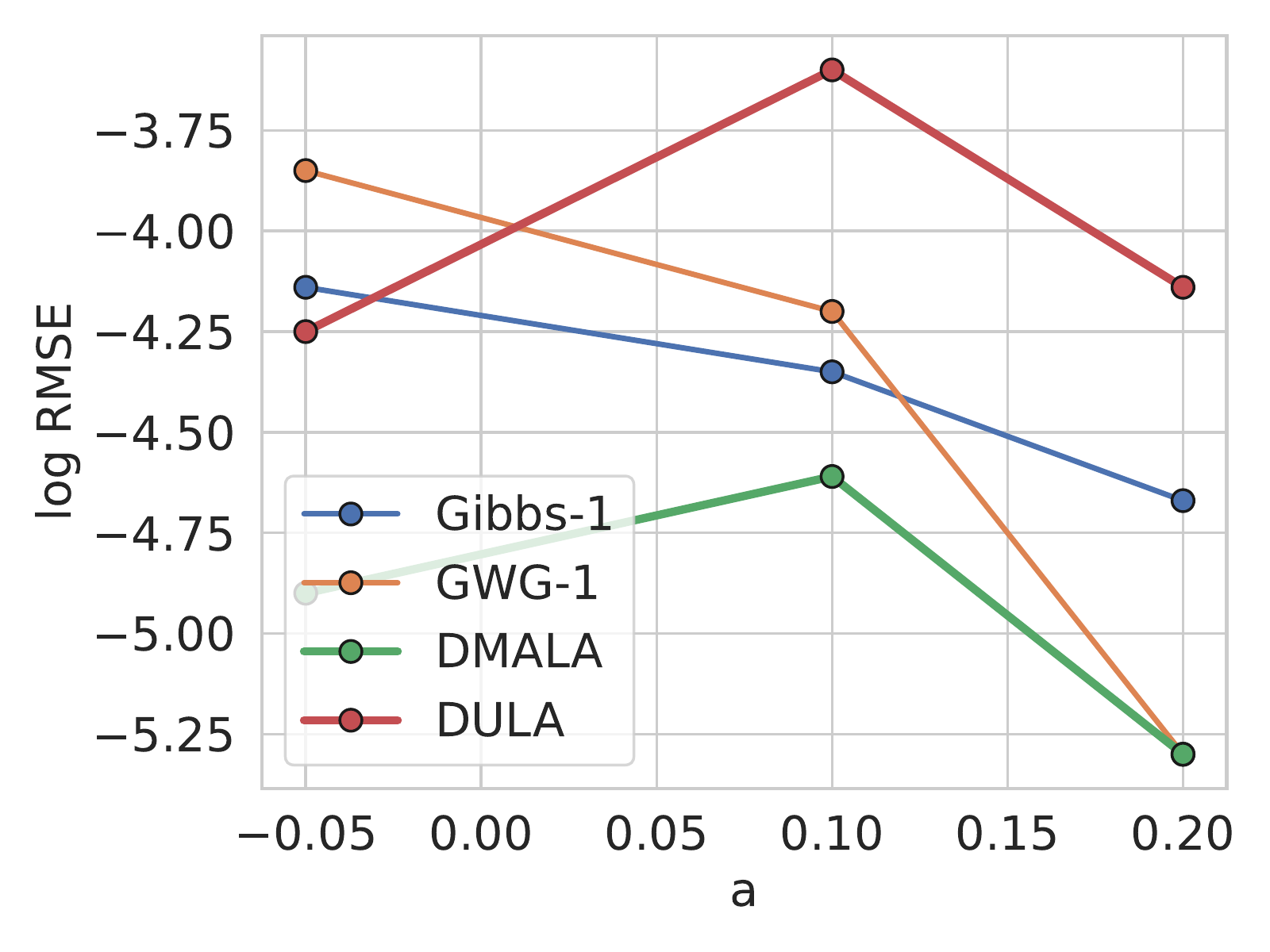} 
	\end{tabular}
	\caption{Ising models with varying connectivity strength}
	\label{fig:ising-sample-sigma}
\end{figure}

We adopt the experiment setting in Section~F.1 of ~\citet{grathwohl2021oops}. The energy of the ising model is defined as,
\begin{equation*}
    \log p(x) = a x^{\intercal} J x + b^{\intercal} x,
\end{equation*}
where $a$ controls the connectivity strength and $J$ is an adjacency matrix whose elements are either $0$ or $1$. In our experiments, $J$ is the adjacency matrix of a 2D lattice graph. DMALA and DULA use a stepsize $\alpha$ of $0.4$ and $0.2$ respectively.
We show the results with varying connectivity strength $a$. We run all methods for the same amount of time (GWG-1 and DMALA for $50,000$ iterations; Gibbs-1 and DULA for $125,000$ iterations). The results are shown in Figure~\ref{fig:ising-sample-sigma}. We can clearly observe that DMALA outperforms other methods on all connectivity strength.

\subsection{Sampling from RBMs}\label{sec:appendix-rbm}
\begin{figure}[H]
\centering
	\begin{tabular}{ccc}		
		\includegraphics[width=0.3\textwidth]{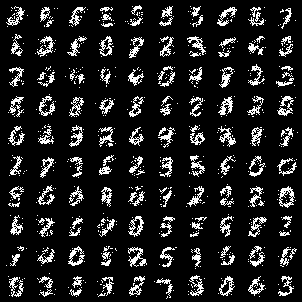}  &
        \includegraphics[width=0.3\textwidth]{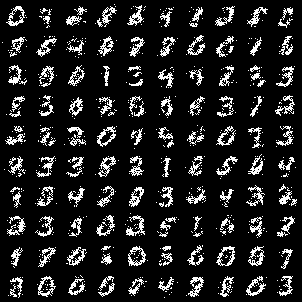}  &
        \includegraphics[width=0.3\textwidth]{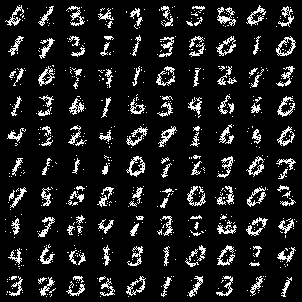}
		\\
		(a) Gibbs-1 &
		(b) HB-10-1&
		(c) GWG-1
		\\
		\includegraphics[width=0.3\textwidth]{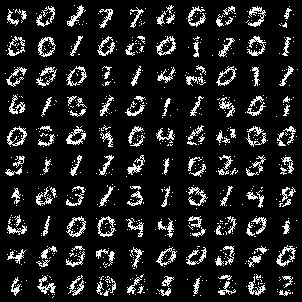}& 
		\includegraphics[width=0.3\textwidth]{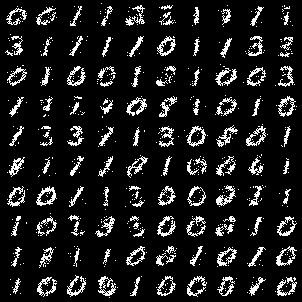}&
		\includegraphics[width=0.3\textwidth]{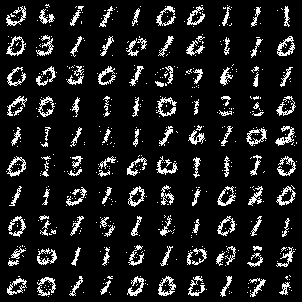}
		\\
		(d) DULA&
		(e) DMALA&
		(f) Block-Gibbs
	\end{tabular}
	\caption{Generated samples from an RBM trained on MNIST. Our methods (DULA and DMALA) generate images that are closer to Block-Gibbs samples.}
	\label{fig:sample-mnist}
\end{figure}

The RBM used in our experiments has 500 hidden units and 784 visible units. The model is trained using contrastive divergence~\cite{hinton2002training}. We set the batch size to 100. The stepsize $\alpha$ is set to be $0.1$ for DULA and $0.2$ for DMALA, respectively. The model is optimized by an Adam optimizer with a learning rate of $0.001$. 
The Block-Gibbs sampler is described in details by ~\citet{grathwohl2021oops}.
Here, we give a brief introduction.
RBM defines a distribution over data $x$ and hidden variables $h$ by,
\begin{equation*}
    \log p(x,h) = h^{\intercal} W x + b^{\intercal} x + c^{\intercal} h  - \log Z.
\end{equation*}
By marginalizing out $h$, we get,
\begin{equation*}
    \log p(x) = \sum_i\text{Softplus}( W x + a)_i + b^{\intercal} x - \log Z.
\end{equation*}
The joint and marginal distribution are both unnormalized and hard to sample from. However, the special structure of the problem leads to easy conditional distributions,
\begin{equation*}
    \begin{aligned}
    p(x|h) &= \text{Bernoulli}(Wx + c) \\
    p(h|x) &= \text{Bernoulli}(W^{\intercal} h+b).
    \end{aligned}
\end{equation*}
Therefore, we can perform Block-Gibbs sampling by updating $x$ and $h$ alternatively. This Block-Gibbs sampler can update all the $784$ coordinates at the same time, and has much higher efficiency. However, this is due to the special structure of the problem. 
We provide the generated images in Figure~\ref{fig:sample-mnist}, showing that the results of DULA and DMALA are closer to that of Block-Gibbs.

\subsection{Learning EBMs}\label{sec:appendix-learned-ebm}
For all the experiments in this section, we set the stepsize $\alpha$ to be $0.1$ for DULA and $0.15$ for DMALA.
\paragraph{Ising Model}
We construct a training dataset of $2,000$ instances by running $1,000,000$ steps of Gibbs sampling for each instance. The model is trained by Persistent Contrastive Divergence~\cite{tieleman2008training} with a buffer size of 256 samples. We also use the Adam optimizer with a learning rate of $0.001$. The batch size is $256$. We train all models with an $\ell_1$ penalty with a penalty coefficient of $0.01$ to encourage sparsity. The experiment setting is basically the same as Section~F.2 in~\citet{grathwohl2021oops}.  

\paragraph{Deep EBMs}
We adopt the same ResNet structure and experiment protocol as in ~\citet{grathwohl2021oops}, where the network has 8 residual blocks with 64 feature maps. There are  2 convolutional layers for each residual block. The network uses Swish activation function~\cite{ramachandran2017searching}. For static/dynamic MNIST and Omniglot, we use a replay buffer with $10,000$ samples. For Caltech, we use a replay buffer with $1,000$ samples. We evaluate the models every 5,000 iterations by running AIS for $10,000$ steps. The reported results are from the model which performs the best on the validation set. The final reported numbers are generated by running $300,000$ iterations of AIS. All the models are trained with Adam~\cite{kingma2014adam} with a learning rate of $0.0001$ for $50,000$ iterations. 

\begin{figure}[H]
\centering
	\begin{tabular}{cccc}		
		\includegraphics[width=0.22\textwidth]{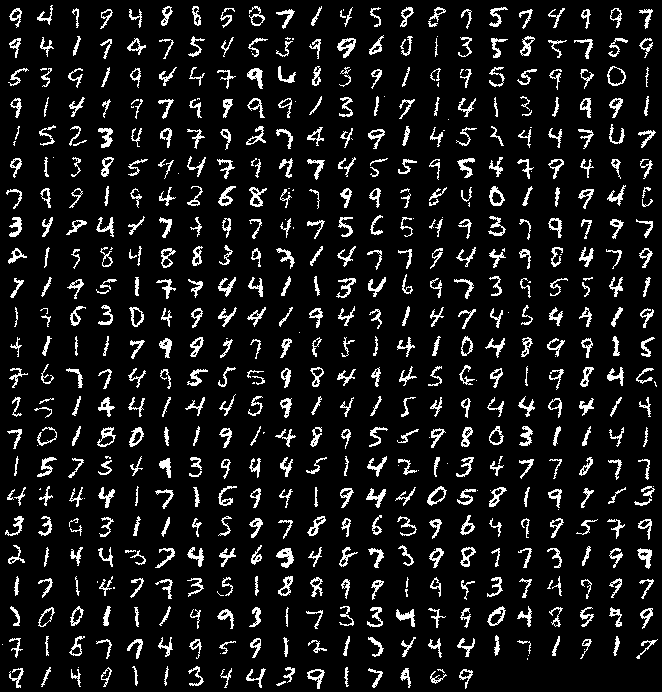}  &
        \includegraphics[width=0.22\textwidth]{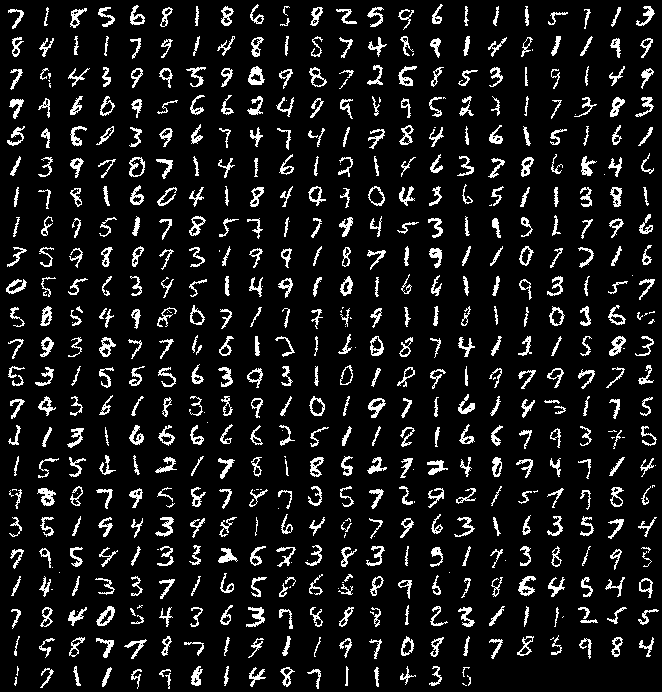}  &
        \includegraphics[width=0.22\textwidth]{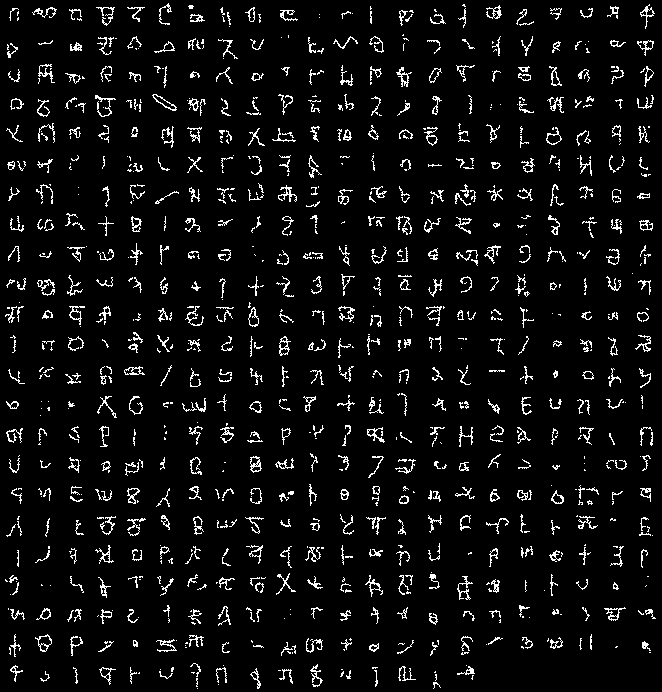}&
        \includegraphics[width=0.22\textwidth]{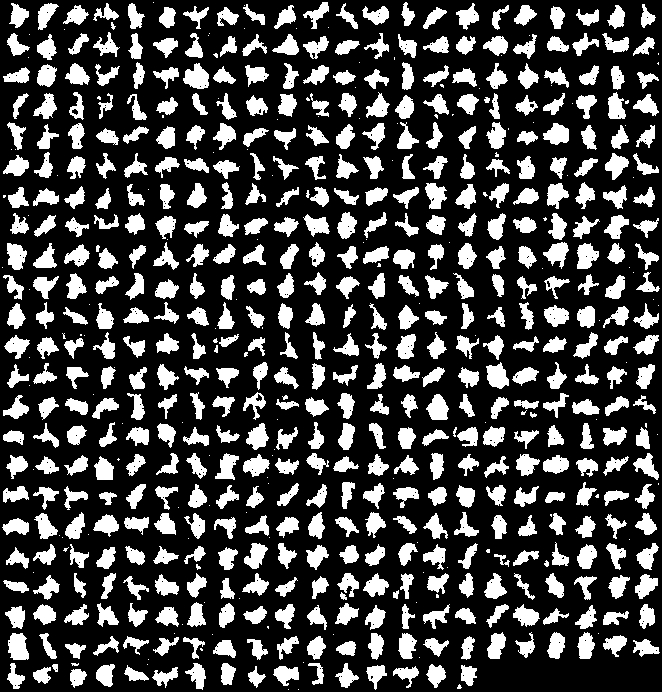}
		\\
		\includegraphics[width=0.22\textwidth]{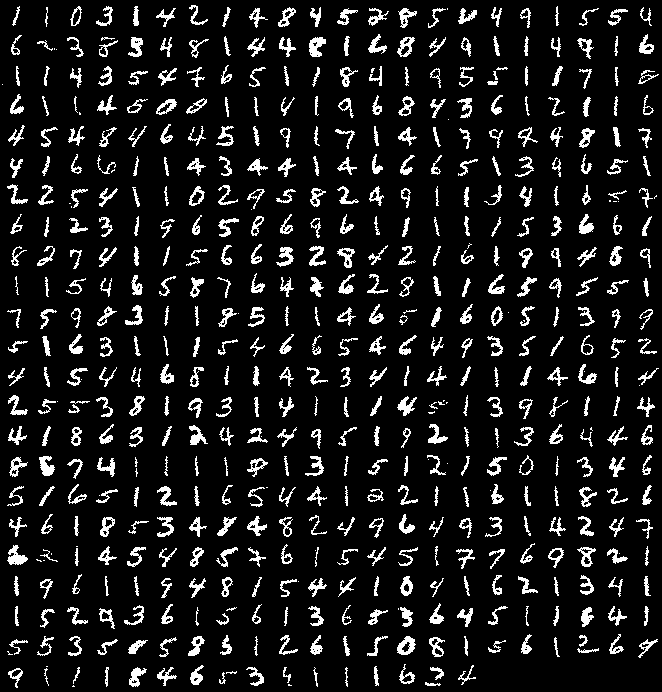}& 
		\includegraphics[width=0.22\textwidth]{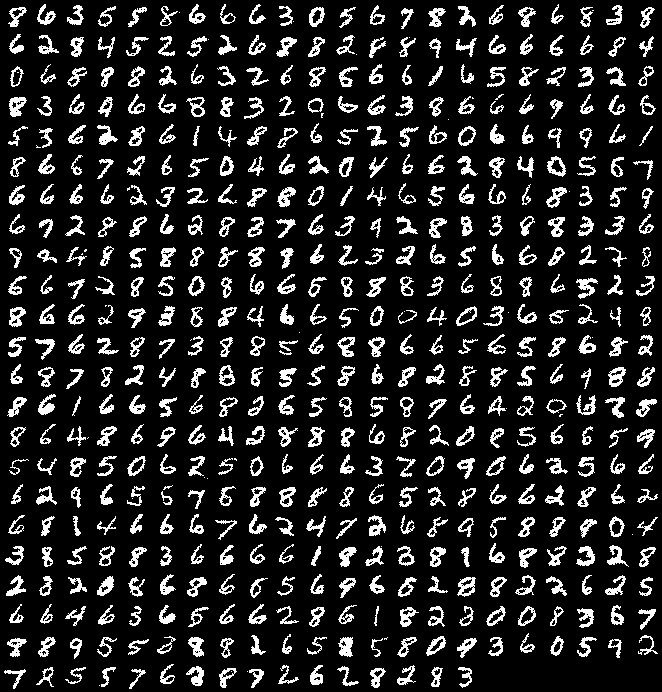}&
		\includegraphics[width=0.22\textwidth]{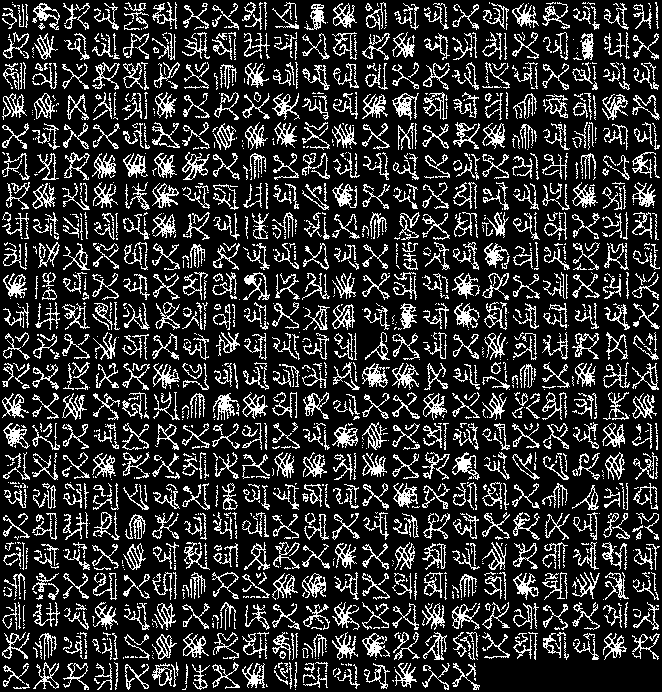}&
		\includegraphics[width=0.22\textwidth]{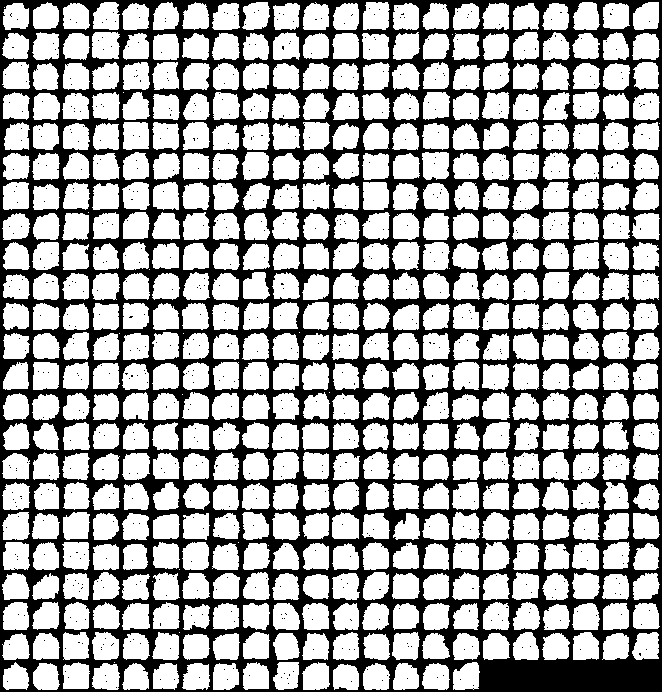}
		\\
	\end{tabular}
	\caption{Left to right: Static MNIST, Dynamic MNIST, Omniglot, Caltech Silhouettes. Top: DMALA. Bottom: DULA.}
	\label{fig:sample-ebm}
\end{figure}

We show the generated images with DULA and DMALA in Figure~\ref{fig:sample-ebm}. We see that the generated images from DMALA are very close to the true images on all four datasets. DULA can generate high-quality images on static and dynamic MNIST, but mediocre images on Omniglot and Caltech Silhouette. We hypothesis that we need to run DULA for more steps per iteration to get better results.

\subsection{Binary Bayesian Neural Networks}
\label{app:bnn}
\textbf{Details of the Datasets} (1) COMPAS: COMPAS~\cite{compas} is a dataset containing the criminal records of 6,172 individuals arrested in Florida. The task is to predict whether the individual will commit a crime again in 2 years. We use 13 attributes for prediction. 
(2) News: Online News Popularity Data Set~\footnote{\url{https://archive.ics.uci.edu/ml/datasets/online+news+popularity}} contains 39,797 instances of the statistics on the articles published by a website. The task is to predict the number of clicks in several hours after the articles published.
(3) Adult: Adult Income Dataset~\footnote{\url{https://archive.ics.uci.edu/ml/datasets/adult}} is a dataset containing the information of US individuals from 1994 census. The prediction task is to predict whether an individual makes more than 50K dollars per year. The dataset contains 44,920 data points.
(4) Blog: Blog Feedback~\citep{buza2014feedback} is a dataset containing 54,270 data points from blog posts. The raw HTML-documents of the blog posts were crawled and processed. The prediction task associated with the data is the prediction of the number of comments in the upcoming 24 hours. The feature of the dataset has 276 dimensions.

\textbf{More Details on Training}
We run 50 chains in parallel and collect the samples at the end of training. All datasets are randomly partitioned into 80\% for training and 20\% for testing. The features and the predictive targets are normalized to $(0, 1)$.  We set $\alpha=0.1$ for all datasets.  We use a uniform prior over the weights. We train the Bayesian neural network for $1,000$ steps. We use full-batch training for the results in Section~\ref{sec:exps} so that Gibbs and GWG are also applicable.

\textbf{Experimental Results of Stochastic DLP} As mentioned in Section~\ref{sec:variants}, we empirically evaluate the performance of stochastic DLP here. We use a batch size of 10, 100, 500 and report the performance of the obtained binary Bayesian neural networks in Table~\ref{tab:bayesbnn_bs}. Note that Gibbs-based sampling techniques are not suitable for mini-batches. We find that our method works quite well with mini-batches, and the performance increases as the batch size becomes larger.

\begin{table*}[ht]
    \centering
    \resizebox{0.99\textwidth}{!}{
    \renewcommand\arraystretch{1.2}{
    \begin{tabular}{c|cccc|cccc}
    \hline \hline
        \multirow{2}{*}{Dataset}&  \multicolumn{4}{c|}{Training Log-likelihood ($\uparrow$)} & \multicolumn{4}{c}{Test RMSE ($\downarrow$)} \\ \cline{2-9}
        & batchsize=10 & batchsize=100 & batchsize=500 & full-batch(DULA) & batchsize=10 & batchsize=100 & batchsize=500 & full-batch(DULA) \\ \hline
        COMPAS & -0.4288 \fs{0.0052} & -0.3689 \fs{0.0027} & -0.3603 \fs{0.0007} & -0.3590 \fs{0.0019} & 0.4929 \fs{0.0052} & 0.4864 \fs{0.0037} & 0.4848 \fs{0.0015} & 0.4879 \fs{0.0024}  \\
        News & -0.2355 \fs{0.0002} & -0.2342 \fs{0.0001} &   -0.2344 \fs{0.0006} & -0.2339 \fs{0.0003} & 0.1072 \fs{0.0013} & 0.0971 \fs{0.0026} & 0.0976 \fs{0.0009} &0.0971 \fs{0.0012} \\
        Adult & -0.3738 \fs{0.0039} & -0.3275 \fs{0.0006} & -0.3233 \fs{0.0009} & -0.3287 \fs{0.0027} & 0.4309 \fs{0.0027} & 0.3963 \fs{0.0016} & 0.3954 \fs{0.0015} &0.3971 \fs{0.0014}\\
        Blog & -0.3199 \fs{0.0072} & -0.2710 \fs{0.0009} & -0.2689 \fs{0.0024} & -0.2694 \fs{0.0025} & 0.3533 \fs{0.0043} & 0.3202 \fs{0.0018} & 0.3198 \fs{0.0026} & 0.3193 \fs{0.0021} \\
    \hline \hline
    \end{tabular}}}
    \caption{Experimental results of binary Bayesian neural networks on different datasets. We examine the influence of batch size on the stochastic version of DLP.}
    \label{tab:bayesbnn_bs}
\end{table*}

\subsection{Text Infilling}
\begin{figure*}[h]
\centering
    \includegraphics[width=0.98\textwidth]{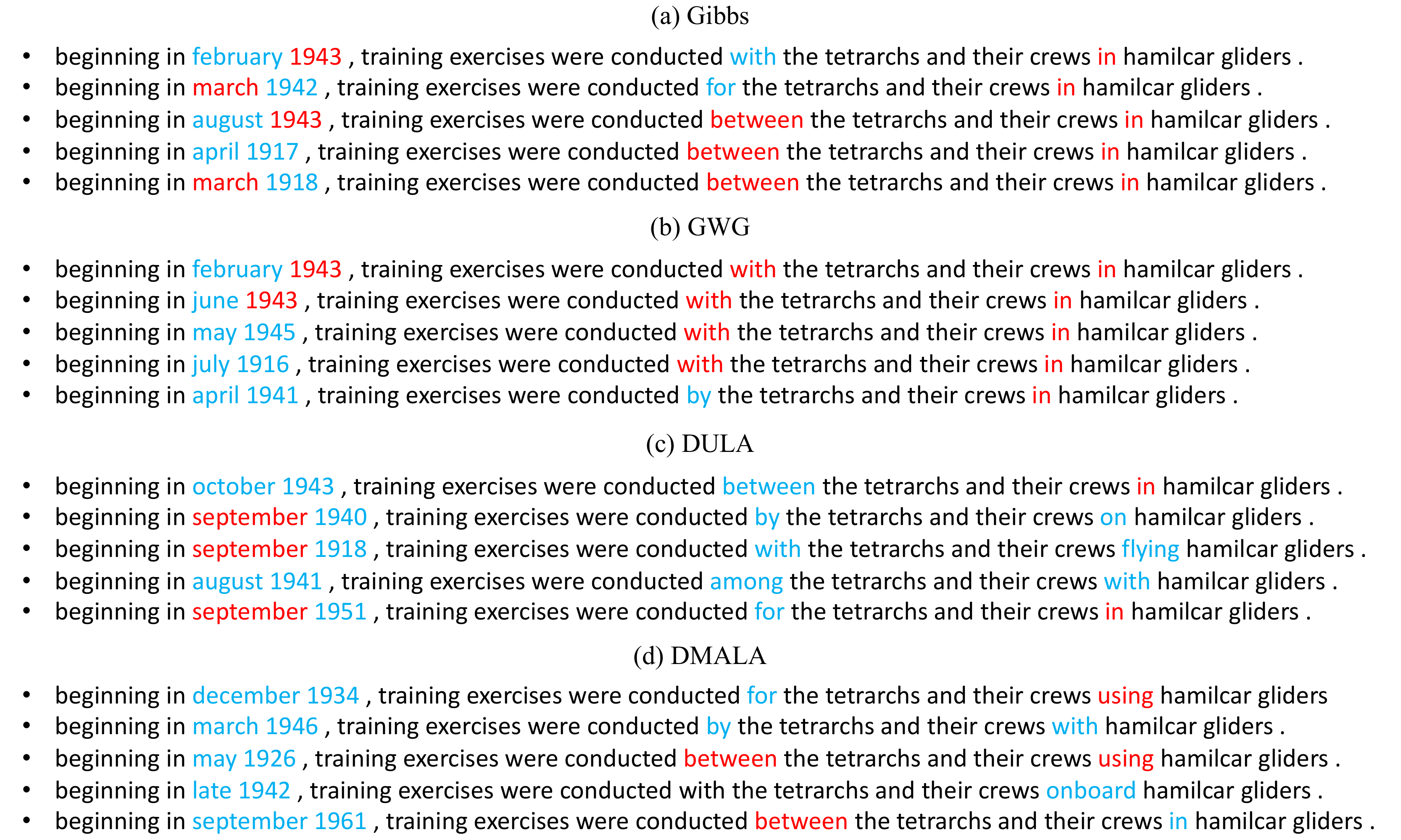}
    \caption{Examples of the generated sentences in the text infilling task. {\color{cyan} Blue} and \red{Red} words are generated. \red{Red} indicates repetitive generation.}
    \label{fig:generate_appendix}
\end{figure*}

We use the BERT model pre-trained in \texttt{PyTorch-Pretrained-BERT}~\footnote{\url{https://github.com/maknotavailable/pytorch-pretrained-BERT}}. 
We use the same TBC and WikiText-103 corpus as in ~\citet{wang2019bert} by randomly sampling $5,000$ sentences from the complete TBC and WikiText-103 corpus. 
BERT is a masked language model, which means that given a sentence with \texttt{MASK} tokens and contexts, the model is able to provide a probability distribution over all the candidate words to indicate which word the \texttt{MASK} position is likely to be. Supposing we have a sentence $X =\left(x_1, x_2, \dots, x_n \right)$ which has $n$ words, and $m$ of the $n$ words are \texttt{MASK}s. 
For each \texttt{MASK[i]}, the BERT model can output a vector $f(\cdot | X_{\text{MASK}[i]}) \in \RR^{N}$ which is a normalized probability distribution over the candidate words representing the probability of a certain word appears at \texttt{MASK[i]} when the context is given.
The BERT model we used has $N=30,522$ candidate words. Hence, $\theta \in \{1,2,\dots,N\}^m$.
Given the user-defined context and \texttt{MASK} positions, we set \texttt{MASK[i]} to $\theta_i$, and
the energy function is defined as,
\begin{equation*}
U(\theta) = \sum_{i=1}^m \log f(\theta_i | X_{\text{MASK}[i]}).
\end{equation*}
There are $N^m$ combinations in total, so directly sampling from the categorical distribution is intractable for $m\geq 2$.
More generated texts are displayed in Figure~\ref{fig:generate_appendix}.

\end{document}